\documentclass[a4paper,twoside]{report}
\usepackage[a4paper, margin=35mm]{geometry}
\usepackage{graphicx}
\usepackage{bookmark}
\usepackage{array,booktabs}
\usepackage{lmodern}
\usepackage{float}
\usepackage{multirow,bigdelim}
\usepackage{textcomp}
\usepackage{fancyhdr}
\usepackage{amsmath}
\DeclareMathOperator{\sign}{sign}

\newcommand{\pvalue}{\mathop{p{\textnormal{-value}}}}
\usepackage{algorithmic,algorithm}
\usepackage{amssymb}
\usepackage{enumitem}
\usepackage{titlepic}
\usepackage{tocvsec2}
\usepackage[titletoc]{appendix}
\usepackage[dvipsnames]{xcolor}
\usepackage{tocloft}
\usepackage{csquotes}
\usepackage{rotating}
\usepackage{color}
\usepackage{hyperref}
\usepackage[acronym,toc,section=section]{glossaries}
\newenvironment{Acknowledgements}
  {
   \begin{abstract}}
  {\end{abstract}
   \clearpage}
\usepackage[nottoc,notlof,notlot]{tocbibind} 
\floatevery{algorithm}{\setlength\hsize{10cm}}

\renewcommand\thesection{\arabic{section}}

\makeglossaries
\hypersetup{colorlinks=false,linkcolor=blue,citecolor=ForestGreen}

\def\thesection{\arabic{section}}
\newcommand\textline[4][t]{%
  \par\smallskip\noindent\parbox[#1]{.600\textwidth}{\raggedright #2}%
  \parbox[#1]{.333\textwidth}{\centering#3}%
  \parbox[#1]{.333\textwidth}{\raggedleft #4}\par\smallskip%
}
\makeatletter
\renewcommand{\l@figure}{\@dottedtocline{1}{2.5em}{4.3em}}
\makeatother

\begin{document}
\begin{titlepage}
\begin{center}

\includegraphics[scale=0.3]{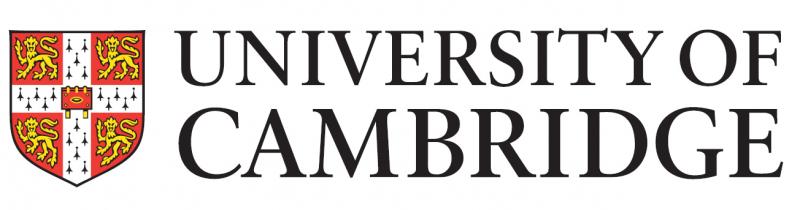}

\vspace{15mm}
\noindent\rule{14cm}{0.4pt}
\vspace{0.1pt}\\
\textbf{\Large A meta-algorithm for classification using random recursive tree ensembles: A high energy physics application}
\vspace{0.1pt}\\
\noindent\rule{14cm}{0.4pt}\\
\vspace{20mm}
\textit{Submitted towards partial fulfilment for the Master of Philosophy in Scientific Computing}\\
\vspace{30mm}
\large{Vidhi Ramesh Lalchand}\\
\vspace{6mm}
\end{center}
\vspace{45mm}
\textline[t]{
\textit{Supervisor:} \\
Dr. Anita Faul \\
Laboratory for Scientific Computing\\
Department of Physics\\
}{}{Clare Hall}
\vspace{10mm}
\begin{center}
August 31, 2016
\end{center}
\end{titlepage}

\thispagestyle{empty}
\textit{This dissertation is substantially my own work and conforms to the University of Cambridge's guidelines on plagiarism. Where reference has been made to other research this is acknowledged in the text and bibliography.}
\clearpage

\thispagestyle{empty}
\null\vfill
\begin{center}
\begin{tabular}{l}
  A man ought to read \\ 
  just as his inclination leads him; \\
  for what he reads as a task \\ 
  will do him little good.\par
 \end{tabular}
 
  \textup{\hspace{35mm} -Samuel Johnson \\\hspace{28mm}1709-1784}
   \end{center} 
\vfill\vfill
\clearpage

\setcounter{secnumdepth}{3}
\setcounter{tocdepth}{3}

\begin{Acknowledgements} 
For this thesis I am indebted to my supervisor Dr. Anita Faul, whose input, advice and above all patience have been very critical during the past year. I would also like to thank Dr. Nikos Nikiforakis for believing in my potential and giving me the chance to pursue this MPhil. Thanks are due to Dr. Christopher Lester for educating me on the wonders of physics, a new field to me. Prof. Alan O'Neil for reading this thesis and guiding its structure and content towards more clarity. Lastly, I would like to thank Zachary Latif for his unwavering belief in my talents, without which none of this would have been possible.
\end{Acknowledgements}


\newacronym{CERN}{CERN}{The European organization for nuclear research}
\newacronym{HEP}{HEP}{High Energy Physics}
\newacronym{ATLAS}{ATLAS}{A Toroidal LHC Apparatus}
\newacronym{SM}{SM}{Standard Model}
\newacronym{AdaBoost}{AdaBoost}{Adaptive boosting algorithm}
\newacronym{AMS}{AMS}{Approximate Median Significance}
\newacronym{NN}{NN}{Neural Networks}
\newacronym{SVM}{SVM}{Support Vector Machines}
\newacronym{DBN}{DBN}{Deep belief network}
\newacronym{BDT}{BDT}{Boosted Decision Trees}
\newacronym{ROC}{ROC}{Receiver Operating Characteristic}
\newacronym{AUC}{AUC}{Area under the ROC curve}
\newacronym{CMS}{CMS}{Compact Muon Solenoid}
\newacronym{LHCb}{LHCb}{Large Hadron Collider beauty}
\newacronym{RF}{RF}{Random Forests}
\newacronym{ET}{ET}{Extremely Random Trees}
\newacronym{BXT}{BXT}{Boosted Extremely Random Trees}
\newacronym{BRF}{BRF}{Boosted Random Forests}

\newglossaryentry{W}{
  name = $W$ ,
  description = W boson
}
\newglossaryentry{Z}{
  name = $Z$ ,
  description = Z boson
}
\newglossaryentry{H}{
  name = $H$ ,
  description = Higgs Boson 
}

\newglossaryentry{photon}{
  name = $\gamma$ ,
  description = Photon 
}

\newglossaryentry{electron}{
  name = $e$ ,
  description = Electron 
}

\newglossaryentry{tquark}{
  name = $b^{+}$ ,
  description = Top quark 
}

\newglossaryentry{bquark}{
  name = $b^{-}$ ,
  description = Bottom quark 
}

\newglossaryentry{tau}{
  name = $\tau$ ,
  description = Tau lepton 
}
\newglossaryentry{neutrinos}{
  name = $\nu$ ,
  description = Neutrino 
}

\begin{abstract}

The aim of this work is to propose a meta-algorithm for automatic classification in the presence of discrete binary classes. Supervised classification at a fundamental level can be defined as the ability to extract rules that discriminate one class from the other. This is done on the basis of training data whose class membership is known with the ultimate objective of classifying new data whose class mapping is unknown. Classifier learning in the presence of overlapping class distributions is a challenging problem in machine learning. Overlapping classes are described by the presence of ambiguous areas in the feature space with a high density of points belonging to both classes. This often occurs in real-world datasets, one such example is numeric data denoting properties of particle decays derived from high-energy accelerators like the \gls{LHCb} at CERN. 
 
A significant body of research targeting the class overlap problem use ensemble classifiers to boost the performance of standard algorithms by using them iteratively in multiple stages or using multiple copies of the same model on different subsets of the input training data. The former is called \textit{boosting} and the latter is called \textit{bagging}. The algorithm proposed in this thesis targets a popular and challenging classification problem in high energy physics - that of improving the statistical significance of the Higgs discovery. The underlying dataset used to train the algorithm is experimental data built from the official ATLAS full-detector simulation with Higgs events (signal) mixed with different background events (background) that closely mimic the statistical properties of the signal generating class overlap. The algorithm proposed is a variant of the classical boosted decision tree which is known to be one of the most successful analysis techniques in experimental physics. The algorithm utilises a unified framework that combines two meta learning techniques - bagging and boosting. The results shows that this combination only works in the presence of a randomization trick in the base learners. The performance of the algorithm is mainly assessed on the basis of a physics inspired significance metric called the \textit{Approximate Median Significance} ($\sigma$). We also show how the algorithm fares compared to the leading machine learning solutions proposed using this dataset.  

\end{abstract}

\tableofcontents 

\clearpage
\listoffigures
\clearpage
\printglossary[type=\acronymtype, title=Abbreviations,toctitle=Abbreviations,nonumberlist]
\printglossary[title=Symbols, toctitle=Symbols,nonumberlist]

\clearpage
\glsunsetall
\addcontentsline{toc}{section}{\listfigurename}

\chapter{Introduction}
\label{intro}

A key property of a particle is how it decays into other particles. \gls{ATLAS}\footnote{A Toroidal LHC Apparatus} is a particle physics experiment at \gls{CERN}\footnote{The European organization for nuclear research} that searches for new particles and processes using head-on collisions of protons at extraordinarily high energies \cite{open}. In 2012, the ATLAS experiment claimed the discovery of a new particle, the Higgs boson. The discovery has a statistical significance of 5$\sigma$ which corresponds to a 1 in 3.5 million chance of the results being obtained purely due to chance. In essence, it denotes a very high confidence in the discovery. The Higgs awaited experimental evidence for over four decades, it was postulated by physicist Peter Higgs in 1964 \cite{higgs}. The existence of this particle provides support to the theory that a field permeates the universe through which fundamental particles acquire mass, a theory which is cardinal for the completeness of the Standard Model of particle physics. The proton-proton collisions in the ATLAS detector produce thousands of collision events per second. Each collision event can be summarised by numeric information represented by a vector of several dimensions. These represent the \textit{features} as in standard machine learning parlance. This thesis views the physics problem of identification of a Higgs decay from a machine learning perspective. The main objective of the thesis is to cast the problem as a binary classification problem and propose an algorithm that addresses the main challenge in the dataset, that of overlapping classes. 

The dataset used to train, cross-validate and test the algorithm proposed in this thesis is obtained from the CERN Open Data portal \cite{data}. The dataset used in this thesis is a modified version of the dataset physicists used in the ATLAS results made public in December 2013 in the CERN Seminar, \textit{ATLAS sees Higgs decay to fermions} \cite{fermions}.

\section{Organization of the thesis}

Chapter \ref{intro} introduces the goal of the problem in a physics context. Chapter \ref{mlhep} provides a discussion of how machine learning techniques are typically used in experimental physics by citing some published approaches. The third chapter \ref{formal} provides a mathematical description of the problem concluding with a summary of the challenges inherent in the machine learning incarnation of the problem. Chapter \ref{formal} also introduces the Approximate Median Significance (\gls{AMS}) metric which is a physics inspired metric used to assess the performance of a binary classifier designed for the task of separating signal from background. It sheds some light on the motivation for the statistical formula of the metric. Chapter \ref{performance} contrasts the AMS metric with standard machine learning metrics in classification tasks. Chapter \ref{ad} introduces the theory behind meta algorithms and Chapter \ref{results} contains performance results of a proposed algorithm on the ATLAS Higgs dataset in terms of the AMS. Chapter \ref{concl} has concluding remarks.

\section{Physics Background} 

Each generation of high energy physics experiments is more demanding in terms of multivariate analysis. Machine learning - known in the physics circles as multivariate analysis played a key role in the Higgs analysis that led to the 2012 discovery. In this section we provide an accessible overview to some of the physics concepts needed to understand the primary data which serve as features to the machine learning model. 

\subsection{Decay Channel}

Particles produced in the proton-proton collisions are unstable and decay almost instantaneously into a cascade of lighter particles. These sets of second order and third order particles represent a \textit{decay} channel or a \textit{decay} product. The surviving particles which live long enough for their properties to be measured by the detector are called \textit{final-state} particles. 

The Higgs boson (H) is unstable and is known to have five main experimentally accessible decay channels. Each occurs which a certain probability, this is called the branching ratio. The branching ratios of the Higgs boson depend on its mass and are precisely predicted in the standard model. The SM predicts branching ratios as a function of the Higgs mass. For a SM Higgs of mass 125 GeV, the first-order decay products and their respective probabilities are :

\begin{center}

\scalebox{0.7}{
\begin{tabular}{|l|l|l|l|}
\hline
\rule[-1ex]{0pt}{2.5ex} Decay Channel & Description & Branching Ratio & Status \\ 
\hline 
\rule[-1ex]{0pt}{2.5ex}  $H \rightarrow b\bar{b}$ & b quark and its anti-quark & 0.577 & predicted \\ 
\rule[-1ex]{0pt}{2.5ex} $ H \rightarrow \tau^{+} \tau^{-} $ & $\tau$ lepton and its anti-particle & 0.063 & predicted \\ 
\rule[-1ex]{0pt}{2.5ex} $ H \rightarrow \gamma\gamma $ & di-photon channel & 0.0023 & observed\\ 
\rule[-1ex]{0pt}{2.5ex} $ H \rightarrow W^{+}W^{-} $ & W boson and its anti-particle & 0.215 & observed\\ 
\rule[-1ex]{0pt}{2.5ex} $ H \rightarrow Z^{0}Z^{0} $ & A pair of Z bosons &  0.026 & observed\\
\rule[-1ex]{0pt}{2.5ex} & Various other decays &   & predicted\\
\hline
\end{tabular}} 
\end{center}
 
The analysis presented in this thesis concerns the $H \rightarrow \tau^{+} \tau^{-}$ channel which characterize the signal events in the dataset. This is explained further in section \ref{taudecay}.

The measured momenta of final state particles are primary information used to identify a Higgs decay.

The ATLAS detector measures three properties of each of the detectable final state particles, they are :

\begin{enumerate}[noitemsep]
\item{The \textit{type} (lepton, hadronic tau, jets)} \item{The \textit{energy}, $E$}
\item{The \textit{3D direction} expressed as a vector $(p_{x}, p_{y}, p_{z})$}
\end{enumerate}

\textit{Note:} Neutrinos are not among the detected final-state particles but appear in the final state. The feature associated with the undetected neutrinos is the \textit{missing transverse momentum}. The concept of transverse momentum deserves a detailed explanation which is provided in section \ref{missing}.

\begin{figure}[ht]
\includegraphics[scale=0.43,width=\textwidth]{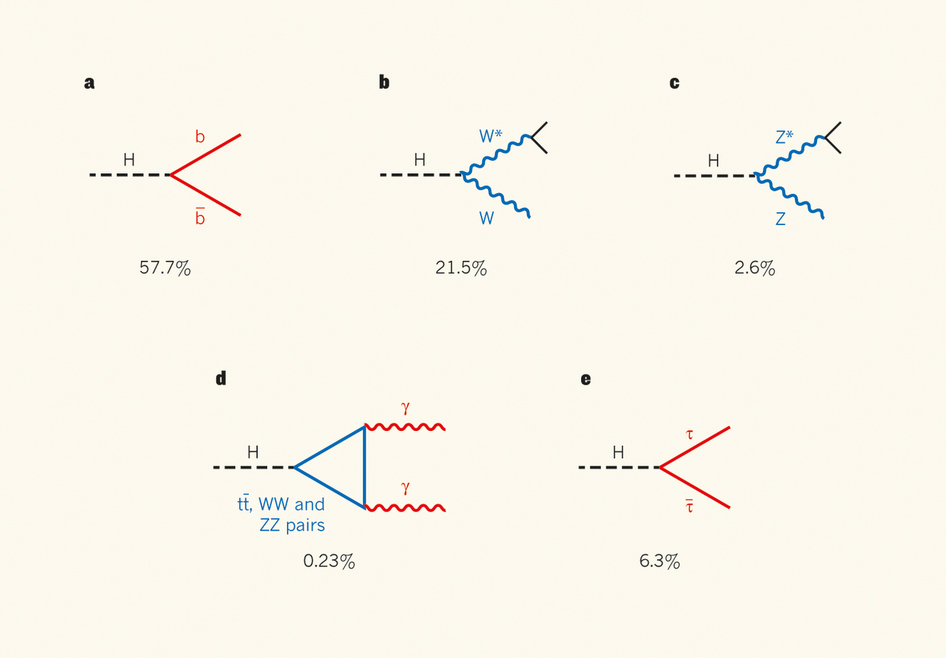}
\caption{Dominant decay modes for the Higgs boson}
\end{figure}

\subsection{Invariant Mass}

The mass of a particle is an intrinsic property of a particle, further by the law of conservation of energy and momentum the mass of a particle is equivalent to the mass of its decayed products each of which can be represented by their 4-momentum ($p_{x},p_{y},p_{z},E$) where $E$ is energy. For example, a particle $\chi$ decays into two final state particles $a$ and $b$ whose kinematics are captured in the detector. By conservation of energy and momentum, $$E_{\chi} = E_{a} + E_{b}$$ $$\overrightarrow{p_{\chi}} = \overrightarrow{p_{a}} + \overrightarrow{p_{b}}$$

The sum of energies and momenta of particles $a$ and $b$ should resolve to give the energy and momenta of the parent particle. The mass of the parent particle is then calculated as, 

\begin{equation}
m_{\chi} = \sqrt{E_{\chi}^2 - \overrightarrow{p_{\chi}}^2} 
\label{mass_invariant}
\end{equation}

Equation \ref{mass_invariant} originates from the energy-momentum relation,

\begin{equation}
E^2 = p^2c^2 + m^2c^4
\end{equation}

relating a particle's intrinsic rest mass $m$, energy $E$ and momentum. In natural units where $c=1$ this simplifies to,

\begin{equation}
E^2 = p^2 + m^2 
\end{equation}

(See Appendix \ref{App})

When a particle decays into lighter particles, its mass before the decay can be calculated from the energies and momenta of the decay products. The inferred mass is independent of the reference frame in which the energies and momenta are measured, so the mass is called \textit{invariant}.

This is the \textit{invariant mass principle} in classical mechanics. It holds for all particles including the Higgs boson and can be generalised to more than two final states and holds in every intermediate stage of decay.   

\subsection{Missing transverse momentum}
\label{missing}

In the 3D reference frame of ATLAS, the $z$-axis points along the horizontal beam line. The $x-y$ plane is perpendicular to the beam axis and is called the \textit{transverse plane}. The transverse momentum is the momentum of an object transverse to the beam axis (or in the transverse plane). The law of conservation momentum promotes the idea of \textit{missing transverse momentum}.

The law of conservation momentum states that the total momentum is conserved in a closed system before and after a collision. We do know that the initial momentum in the plane transverse to the beam axis is zero. Hence, the sum of transverse momentum of all particles (detected + undetected) post-collision should be zero. 

The missing transverse momentum is defined as, $E_{miss}^{T} =  - \sum_{i} \vec{p_{T}}(i) $ for visible particles $i$ where $\vec{p_{T}}$ is the transverse momentum. Essentially, a net momentum of outgoing visible particles indicates missing transverse momentum attributed to particles invisible to the detector, like neutrinos. We know that the final state events consists of neutrinos and it is reasonable to estimate that they make up a lot of the missing transverse momentum.

\subsection{Tau Decay}
\label{taudecay}

In the original discovery the Higgs was seen decaying into $\gamma\gamma$, $W^{+}W^{-}$ and $Z^{0}Z^{0}$. The $H \rightarrow \tau^{+} \tau^{-}$ channel is particularly interesting as it hasn't been experimentally verified .i.e. its statistical significance is not yet at 5$\sigma$. 

It is important to understand what makes this specific decay channel hard to observe. There are two main reasons for this :

\begin{enumerate}

\item The decay into two taus is not a unique channel, in fact the Z boson can also decay into two taus, further this happens a lot more frequently than the Higgs. The precise mass of the Z boson is 91 GeV, since this is not very far from the mass of the target Higgs (125 GeV), the two decays produce events which have very similar signatures and this prevents a clean separation of the parent candidate.

\item Taus are heavy and unstable, they decay instantaneously. Their dominant decay modes involve neutrinos and the presence of these undetectable particles in their decay make it difficult to reconstruct the parent mass on an event by event basis.
\end{enumerate}

The three dominant channels of $\tau$ decay are: 
\begin{enumerate}[noitemsep]
\item{$\tau \rightarrow e^{-}\nu_{e}\nu_{e}$} [an electron and two neutrinos]
\item{$\tau \rightarrow \mu^{-}\nu_{\mu}\nu_{\mu}$} [a muon and two neutrinos]
\item{$\tau \rightarrow$  $\tau$-hadron and $\nu_{\tau}$ [a tau-hadron and a neutrino]}
\end{enumerate} 
The data underlying the results in this thesis focuses on the $H \rightarrow \tau^{+} \tau^{-}$ decay channel where the signal events indicate a Higgs decay to two taus and background events are characterized by the same tau-tau channel but from the decay of a non-Higgs particle.

\begin{figure}
\includegraphics[width=\textwidth]{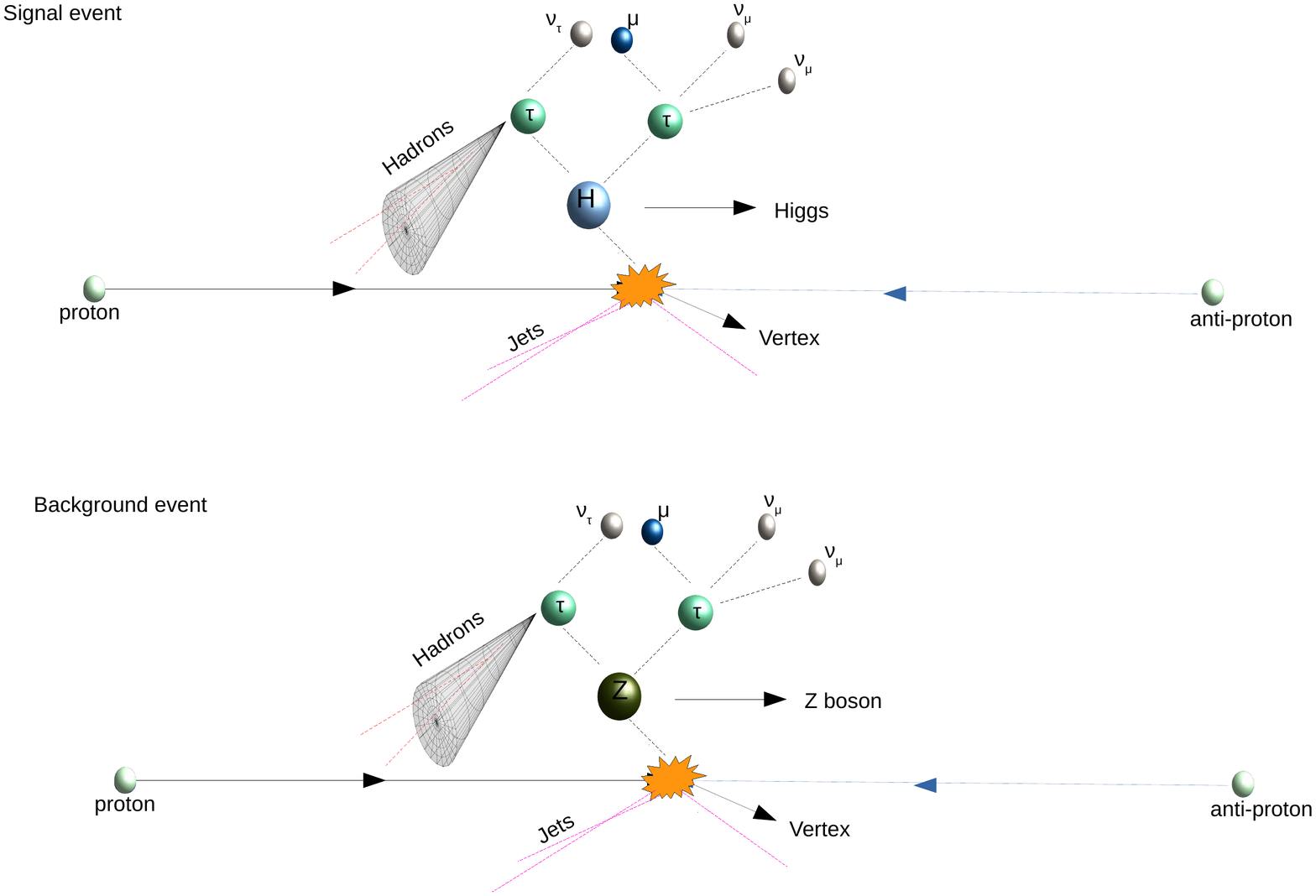}
\caption{H and Z decay channel with similar signature}
\end{figure}

\subsection{Collider events}

The LHC collides bunches of protons every 50 nanoseconds at four interaction points. Each bunch crossing yields more than 20 proton-proton collisions on average. The average number of collisions per bunch crossing is between 10 and 35 depending on the conditions inside the collider. The colliding protons produce a small firework in which new particles are released as a result of the kinetic energy of the protons. An online trigger classifier system discards a vast majority of bunch collisions which contain uninteresting events, this decreases the event rate from 20,000,000 to about 400 per second. The selected 400 events are saved on disk producing about 1 billion events and three petabytes of raw data per year.   

\subsection{Simulated Data}

The dataset used in this thesis is a simulated dataset constructed by ATLAS physicists. Because the problem is the discovery of new phenomenon, labelled samples of actual signal events are not available. Hence, classifiers cannot be trained on real accelerator data. Instead, the data are drawn from elaborate simulators of the accelerator which generate events following the rules of the Standard Model and take into account noise and other artefacts. The simulators are sophisticated models that capture the best current understanding of the physical processes and have been thoroughly verified using data from previous accelerators. The classifiers are trained and validated on such data before they are applied to real unseen data with no class labels.   

\subsection{Experimental search process}
\label{esp}

The Higgs ($H$) is unstable and decays almost instantaneously into lighter particles, further its occurrence is rare. In order to create conditions for a $H$ decay two beams of protons are accelerated to energies close to the speed of light and collided inside a particle detector. The detector cannot directly observe $H$ but registers properties of the decay products. The reconstructed decay may match a possible $H$ decay but this is not enough to establish if $H$ was actually created. Many parent particles could have produced similar decay signatures. This complicates direct analytical inference. However, the SM predicts the likelihood of decay signatures of each know process. Hence, if the detector detects more decay signatures consistently matching a $H$ decay than would have been expected if $H$ did not exist, then that would be strong evidence that $H$ does exist. More precisely, the excess has to be atleast 5$\sigma$ i.e., the observed decays need to be more than 5 standard deviations away from the expectation if there was no new particle, in this case no Higgs.  

The question is really that of statistical significance, because the occurrence of $H$ is so rare and a high threshold of statistical significance needs to be reached a large number of collision events need to be analysed to ensure that correct conclusions are being drawn.  

The 4th July, 2012 announcement claiming the Higgs discovery under the di-photon channel entailed sifting through over 300 trillion ($3$x$10^{4}$) proton-proton collisions using the world's largest computing grid at the Large Hadron Collider at CERN.

\begin{sidewaysfigure}
\includegraphics[width=\textwidth]{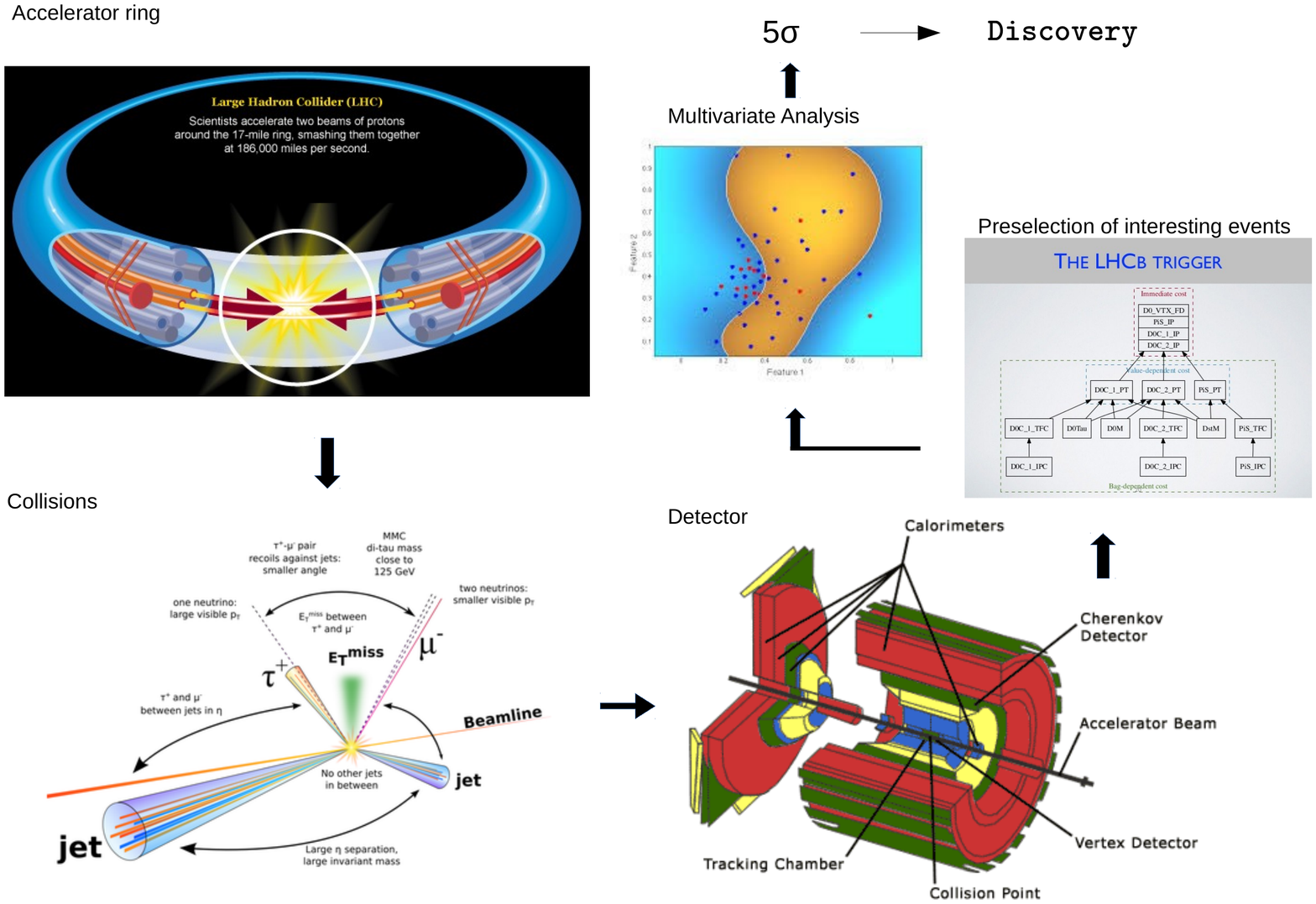}
\caption{Illustration of experimental search}
\end{sidewaysfigure}

\chapter{Machine learning in HEP: A review}
\label{mlhep}

\section{Supervised learning for event classification}

Classifying new particles is not quite like generic classification tasks\footnote{Classifying cat videos would be an example of a generic classification task.}. The main reason is that in experimental high energy physics (HEP) we are looking to verify whether events of a certain class exist or not. More formally, we are looking to verify one of two hypotheses,

$H_{1}$ : A certain decay/event is known to happen \\
$H_{0}$ : It doesn't 

The rejection of the null hypothesis entails in detecting an excess of events of the positive signal class with a high significance level as described in Chapter 1 section \ref{esp}. The detection of the excess occurs through casting this question into a supervised binary classification framework where each event is classified into one of two classes - signal or background. Most problems that harness supervised classification algorithms derive conclusions from evaluating the direct accuracy of the classification process on unseen data. In HEP, the important factor is not accuracy per se but significance. In this respect, HEP differs from other standard applications of machine learning classification. 

At the heart of most classification models lies a mathematical function that calculates a value $f(x)$ on an input $x$. This can be thought of as a discriminant score taking small values for the  negative class (background) and large values for the positive class (signal). The classification step assigns to each event $x$ the associated discriminant value $f(x)$. The discriminant values can be thought of as a ranking of the events where the events most likely to be signal have larger values and vice versa. By applying a threshold on the discriminant score a selection region is derived where an event with a discriminant value larger than the threshold, say $f(x) > \theta$ is predicted to be a signal event and the others are predicted to be background events. 

The selection region is a (not necessarily connected) region in the feature space where an excess of signal events is expected over background. It is fully determined by the discriminant value function $f(x)$ and a choice of threshold $\theta$. It consists of all events which are predicted to be signals by the classifier.

The acceptance or rejection of the null hypothesis is based on the number of true positives and false positives which lie in the selection region of a classifier. The true positives in the selection region are events with a true class label of signal (these are the hits or correct predictions by the classifier) and false positives are events in the selection region with a true class label of background.

From a machine learning perspective we are interested in building a classifier which yields a selection region that has an excess of true positives over false positives. The degree of this excess is directly connected to the significance measurement.

The motivation for the section above is to give a flavour of the steps entailed in event classification in the context of supervised learning. The formal framework of the steps described above and the measurement of discovery significance are described in Chapters \ref{formal}, \ref{performance}.

The sections below discuss published approaches of some of the most popular machine learning techniques for event classification at experiments at CERN and at other particle colliders. 

\section{Neural Networks (NN)}

NNs had a timid start in high energy physics in the late 80s but since the last decade and the advent of deep learning, NNs have started to be used more broadly. This is probably both due to an increased complexity of the data to be analysed and the demand for non-linear techniques. 

NNs are used for both trigger classifiers and off-line data analysis. Trigger classifiers are used as online algorithms that discard uninteresting\footnote{Events which are unlikely to be of the positive signal class.} events as and when they are generated. Online trigger systems at the LHC reduce the rate of data-collection by several orders of magnitude and have sub-millisecond response times. The trigger systems frequently use specialized neural network chips for the task. Two experiments which use neural network triggers are DIRAC \cite{dirac} and H1 \cite{h1}. In the H1 trigger system, a multi-stage system is used with a feed-forward 3 layer NN applied in each stage with the output being 1 or 0 for signal or background respectively. 

For particle identification in offline data analysis, the most successful application of NN has been in detecting the decay of the Z boson. A feed-forward network was used to discriminate the decay of the Z boson into c, b or s quarks. 

One of the main advantages of NNs is that they are intrinsically parallel and tolerant to noise. However, the poor interpretability of NN outputs versus other techniques like Boosted Decision Trees (BDTs) are marked disadvantages. Most of the current work involving applications of NNs to HEP are in the space of deep learning, section \ref{deep} includes a discussion.  
 
\section{Boosted Decision Trees (BDT)}

BDTs have been by far the most popular technique for analysing data from HEP experiments for particle identification. In the MiniBooNE experiment \cite{yang} at the Fermi National Laboratory (Fermilab) for neutrio oscillations, tuned BDTs were used after a careful comparison of several boosting algorithms. The study shows that the BDT is not only better at event separation but is also more stable and robust than NN. In BDTs, decision trees act as component classifiers, the component classifiers are also called `weak' learners which are applied stage-wise. At each stage events that are misclassified are over-weighted and the same weak classifier is applied on the re-weighted data. This process continues until the error metric saturates. AdaBoost is known to be the most successful algorithm based on the boosting technique, \cite{yang} confirms on the basis of numerous trials that AdaBoost was superior in performance to the rest. 

\cite{cascade} uses a cascade idea with a two stage training process involving either a BDT or a NN or both in individual stages. The intuitive idea is to show that successive training improves the performance of a single stage classifier and the best results are obtained through hybridization .i.e. when using a different classifier at each stage.   

\section{Deep learning in HEP}

\label{deep}

In many of the problems in experimental physics we don't know what we are looking for, events are not labelled, this creates a classic use case for deep learning which fundamentally works as an unsupervised process attempting to model high level abstractions in the data. 

The big question for the future is if deep learning techniques can outperform the current learning methodologies and significantly improve discovery significance.

There are two open questions surrounding this debate. 

\begin{enumerate}
\item Can deep learning techniques beat the abilities of veteran physicists in coming up with features that have high discriminatory power? Can they do it just by looking at the raw data from the collider? 
\item Deep learning is known to be computationally expensive, does the marginal contribution to classification power over relatively shallower learning architectures justify the computational cost?
\end{enumerate}

There is a serious computational challenge in training models with deep architectures that usually involve several layers of adaptive parameters.  

In \cite{deep} they use a deep network architecture for the $H \rightarrow \tau^{+} \tau^{-}$ benchmark search and claim to improve the AUC score (area under the ROC curve, see Chapter \ref{performance}) by 8\% and achieve a discovery significance of 5$\sigma$. They divide the feature set into high-level and low-level features. The low-level features are the raw quantities captured by the particle detector like type, energy and momentum of the particles in the decay product. High-level features on the other hand refer to physical quantities that are computed using the low-level features. The shallow methods trained with only low-level features perform worse than with only high-level features implying that shallow methods are not able to independently discover the information reflected in high level features. This motivates the calculation of high-level features. While methods trained only on high-level features perform worse than those trained on the full suite of features, deep architectures show nearly equivalent performance using low-level features and complete feature set suggesting that they are automatically discovering the insight contained in high-level features. These results demonstrate the advantage of using deep learning techniques relative to current approaches. The data used for this study did not come from the official ATLAS event simulator and the author does not discuss the computational set-up for the experiment.  

Owing to the popularity and wide acceptability of BDTs in HEP, the algorithm proposed in this thesis is a variant on the traditional boosted machine. The next chapter introduces the formal $H \rightarrow \tau^{+} \tau^{-}$ problem in a classification context.

\chapter{The formal problem}
\label{formal}
\section{Data Semantics}

The particles and pseudo-particles that appear in the final state of the collision events in the dataset are :

\begin{enumerate}[noitemsep]
\item{Hadronic-tau} 
\item{Lepton} 
\item{Leading Jet}
\item{Sub-leading Jet}
\end{enumerate}

The primary features in the dataset comprise of 3 measured properties of each of the detectable final-state particles and pseudo-particles. The measured properties are:

\begin{itemize}[noitemsep]
\item Pseudorapidity 
\item{Azhimuth angle} 
\item{Transverse momentum}
\end{itemize}

Apart from the features each collision event in the data has an additional attribute - \textit{weight}. In the real data, the classes are very imbalanced, the probability of a signal event in the natural world  is several magnitudes lower than that of a background event. However, the dataset used for the analysis has been enriched with signal events to generate a more balanced classification problem. To compensate for this bias, all events are weighted with importance weights reflecting their true probability of occurrence. The weight of each event is a non-negative quantity which corrects for the mismatch between the natural probability of a signal event and the probability applied by the simulator. The mathematical meaning behind the importance weights is dealt with in section \ref{math} of this chapter. The importance weights are not meant to be given as inputs to the classifier as the weight distribution of the signal and background events are very different and this would give away the class label immediately. The ratio of signal to background events in the data is roughly 30:70. While the weights are not used as inputs, they are used to assess the performance of the classifiers. 

\section{Features}
\label{features}

Below we include a brief description of the physical meaning behind the features provided in the dataset for each particle in the decay product. 

In the 3-d reference frame, we assume the $z$-axis to be the horizontal beam line. Transverse quantities are quantities projected on the plane perpendicular to the beam line, this is the $xy$ plane. The primary ingredients needed to compute the characteristics of the parent particle are the 4-momentum vectors $(p_{x}, p_{y}, p_{z}, E)$ for the particles in the decay products. The primary features in our dataset are computed from the raw 4-momentum coordinates. These physical quantities constructed by ATLAS physicists capture properties of the decay channel most critical to the inference of the parent particle. Below we describe these quantities which are used as features in our problem. The dataset comprises these quantities for each particle in the final-state of the collision \cite{rm}.

\begin{figure}[ht]
\begin{center}
\includegraphics[scale=0.7]{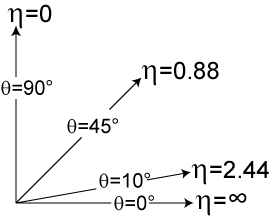}
\caption{Understanding pseudorapidity feature}
\label{pseudo}
\end{center}
\end{figure}

\textbf{Pseudorapidity ($\eta$)} : This describes the angle of the particle relative to the beam axis. It is defined as, $$ \eta = -\ln [\tan(\theta/2)]$$ where $\theta$ denotes the angle between the particle and the positive direction of the beam axis. Figure \ref{pseudo} depicts the concept, 

$\eta = 0 $ corresponds to a particle in the $xy$ plane perpendicular to the beam line, $\eta = +\infty$ corresponds to a particle travelling along the z-axis in the positive direction and $\eta = -\infty$ denotes travel in the opposite direction. Particles with high $\eta$ are usually lost and not captured by the detector. 

Particles can be identified in the range $\eta \in (-2.5 +2.5)$, for $|\eta| \in [2.5,5]$, their momentum can be measured but the particle cannot be identified. Particles with $|\eta| > 5$ escape detection all together \cite{rm}. 

\textbf{Azimuth Angle ($\phi$)} : Decay particles shoot out from the vertex of the collision which lies on the $z$-axis. The vector from the vertex to the particle is projected onto the transverse plane ($xy$), the angle between the projected vector and the $x$-axis is the azimuth angle. 

\begin{figure}
\begin{center}
\includegraphics[scale=0.6]{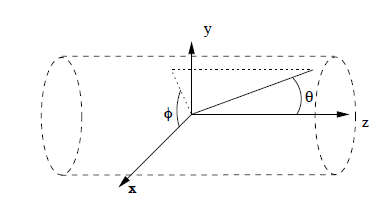}
\caption{Particle collider reference frame}
\end{center}
\end{figure}

\textbf{Transverse momentum  ($p_{t}$)} : The transverse momentum can be defined as the momentum that materializes in the $xy$ plane perpendicular to the beam axis. A hard collision event is characterized by a high $p_{t}$, while proton collisions that result from protons brushing against each other leave decay particles not too far from the beam axis resulting in a small $p_{t}$. 

The transverse momentum is computed as, $$ p_{t} = \sqrt{p_{x}^2 + p_{y}^2}$$
It is possible to derive the momentum vector $\mathbf{p} = (p_{x},p_{y},p_{z})$ from $\phi$, $p_{t}$ and $\eta$.

\begin{equation}
\mathbf{p} =  \left( \begin{aligned} p_{x} \\p_{y} \\p_{z}  \end{aligned} \right) = \left( \begin{aligned} p_{t} \times \cos(\phi) \\ 
									p_{t} \times \sin(\phi) \\ p_{t} \times \sinh(\phi) \end{aligned} \right)
\end{equation}

\section{Mathematical Description}

\label{math}

The description in this section is based on Section 4.1 of \cite{rm}.\\

Let $\mathcal{D} = \{(\mathbf{x}_{1},y_{1},w_{1}),...,(\mathbf{x}_{n},y_{n},w_{n})\}$ be the sample data set provided by ATLAS, $\mathbf{x}_{i} \in \mathbb{R}^d$ is a \textit{d}-dimensional feature vector, $y_{i} \in \{b,s\}$ is the class label and $w_{i} \in \mathbb{R}^{+}$ is a non-negative weight associated with each sample. Let $\mathcal{S} = \{i : y_{i} = s\}$ and $\mathcal{B} = \{i : y_{i} = b\}$ represent index sets of signal and background events respectively. Also, $n_{s} = |\mathcal{S}|$ and $n_{b} = |\mathcal{B}|$ represent the number of signal and background events in the dataset. 

The simulated dataset differs from the real-world dataset in the frequency with which signal events occur. The natural world probability of occurrence of signal events is much lower than what is reflected in the dataset. The ratio of the number of signal events to background events in the dataset $n_{s}/n_{b}$ is not reflective of the true ratio of prior class probabilities $P(y = s)/P(y = b)$, this is because $P(y = s) \ll P(y = b)$ and the true distribution of events in the dataset would yield an extremely unbalanced classification problem with $n_{s}$ significantly lower than $n_{b}$.  The simulated dataset is enriched with signal events to present a more balanced classification problem. In order to correct for this bias, all events are weighted with importance weights. In the dataset the average weight of a signal event is 300 times smaller than the average weight assigned to a background event. 

For each class, the quantities $N_{s}$ and $N_{b}$ are defined as, 
\begin{equation}
\sum_{i \in \mathcal{S}} w_{i} = N_{s} \hspace{5mm} 
\textrm{ and } \hspace{5mm} 
\sum_{i \in \mathcal{B}} w_{i} = N_{b} 
\label{weights}
\end{equation}

These constants have physical meaning, they are the expected total number of signal and background events during the time interval of data taking (in the dataset used, it is the year 2012). The objective function (introduced below and described in section \ref{motivationams}) that the classifier needs to optimize depends upon the quantities $N_s$ and $N_b$ rather than the number of signal and background events $n_s$ and $n_b$. The weights are normalized such that their sum is explicitly set to the expected number of events, if the time interval of data taking is expanded the expected number would change and this would require a re-normalization of the weights.  

The weights $w_{i}$ are defined as follows, 

\begin{equation}
w_{i} \approx \begin{cases} p_{s}(\mathbf{x}_{i})/q_{s}(\mathbf{x}_{i}) \textrm{ if } y_{i} = s, \\ p_{b}(\mathbf{x}_{i})/q_{b}(\mathbf{x}_{i}) \textrm{ if } y_{i} = b, 
\end{cases}
\label{wratio}
\end{equation}

where $p_{s}(\mathbf{x}_{i})$ and $p_{b}(\mathbf{x}_{i})$ are the natural probabilities of occurrence of a signal/background event and $q_{s}(\mathbf{x}_{i})$ and $q_{b}(\mathbf{x}_{i})$ are the probability densities used by the simulator. The weights are just the ratio of the true probability of an event to the simulator applied probability of the event. 

A method or analysis that yields a certain threshold of performance on the given dataset with roughly 1 million events should yield a very similar performance when the dataset is scaled up. This is because we make the set-up invariant to the number of signal and background events by using the sum of importance weights to reflect their true rates of occurrence. 

Further, for ease of analysis the weights have been distributed in such a way that the sum across weights across the training set, test set and cross-validation set are kept fixed. 
Again, this is to make the performance metric based on weights comparable across the different sets which have different number of signal and background events.   

Let $h: \mathbb{R}^{d} \rightarrow \{b,s\} $ be an arbitrary binary classifier. The selection region $\mathcal{H} = \{\mathbf{x} : h(\mathbf{x}) = s\}$, $\mathbf{x} \in \mathbb{R}^{d}$ is the set of points classified by $h$ as a signal, these are the \textit{predicted} positives. Let $\hat{\mathcal{H}}$ denote the index set of points that $h$ classifies as signal, 

\begin{equation}
\hat{\mathcal{H}} = \{i : \mathbf{x}_{i} \in \mathcal{H}\} = \{i : h(\mathbf{x}) = s \} 
\label{index_select}
\end{equation}

The quantities, 
\begin{equation} 
\hat{s} = \sum_{i \in \mathcal{S}\cap \hat{\mathcal{H}}} w_{i} \hspace{5mm}
\textrm{    and    } \hspace{5mm}
\hat{b} = \sum_{i \in \mathcal{B}\cap\hat{\mathcal{H}}} w_{i} 
\label{unbiased}
\end{equation} 

are the true positives and false positives. Here, the weights are used as a proxy for the number of signal and background events. In the real analysis one would just count the number of events in the selection region.  

A typical binary classifier $h: \mathbb{R}^{d} \rightarrow \{b,s\}$ calculates a discriminant function $f(\mathbf{x}) \in \mathbb{R},\mathbf{x} \in \mathbb{R}^{d}$ which is a score giving small values for the negative class (background) and large values for the positive class (signal). One puts a threshold of choice $\theta$ on the discriminant score and classifies all samples below the threshold as belonging to the negative class ($b$ or$-1$) and all samples with a score above the threshold as belonging to the positive class ($s$ or $+1$). 

The discriminant function $f(\mathbf{x})$, also called \textit{decision function}, is evolved at the time of training and applied to test samples to reach classification decisions.

Most classifiers are optimized to improve classification accuracy on a held-out test set. The classification accuracy is the fraction of correctly classified samples belonging to all classes. Using the terminology from the table below, 

\begin{table}[h]
\begin{center}
\begin{tabular}{c|c|c|c}
 & & \multicolumn{2}{c}{Predicted Label}\\
 \hline
 & & -1 (b) & +1 (s) \\
 \hline
\multirow{3}{*}{True Label} & -1 (b) & True Negatives (TN) & False Positives (FP)  \\ 
& +1 (s) & False Negative (FN) & True Positives (TP) \\
\end{tabular}
\label{cf}
\caption{Confusion Matrix}
\end{center}
\end{table}

and the fact that positives (P) = TP + FN and negatives (N) = TN + FP, the classification accuracy is defined as the fraction $\frac{\displaystyle \text{TP + TN}}{\displaystyle \text{P + N}}$. When class distributions are imbalanced, a metric such as the overall classification accuracy is a weak indicator of the performance of a classifier. This is because the class distributions are skewed rather than balanced. Given that around 70\% of the samples belong to the negative class, a classifier that assigns each sample to the negative class will have an accuracy score of 70\%, but this ignores the performance of the classifier with respect to classifying samples of the positive minority class correctly. 

Hence, in many contexts the question surrounding reliable performance measurement is tied to the problem at hand. For instance, in bio-informatics, the significance of a discovery is tied to whether the false discovery rate, defined as, $\frac{\displaystyle \text{FP}}{\displaystyle \text{FP+TP}}$ is small enough. 

In a similar spirit, the physicists at ATLAS specify an objective function to be maximized by the classifier. It is called the \textit{Approximate Median Significance} (AMS) metric. It is sometimes used loosely as discovery significance metric to emphasize the role it plays in the discovery process of new phenomenon. 

Given a binary classifier $h : \mathbb{R}^{d} \rightarrow \{b,s\}$, the AMS is given by, 

\begin{equation}
\textrm{ $AMS_{s}$ } = \sqrt{2((\hat{s} + \hat{b})\ln(1 + \frac{\hat{s}}{\hat{b}})-\hat{s})} 
\label{ams} 
\end{equation}
\raisetag{-.4em}

where $\hat{s}$ and $\hat{b}$ are the expected number of signal and background events as in eq. \ref{unbiased}.

Eq. \ref{ams} shows that the AMS is fully determined by quantities $\hat{s}$ and $\hat{b}$ which are computed using the events in the selection region. 

\section{Motivation for the AMS Objective}
\label{motivationams}

The events in the selection region $\mathcal{H}$ of a classifier belong to one of two categories:

\begin{itemize}
\item{Selected Background events: 
\begin{equation*} 
\hat{b} =\sum_{i \in \mathcal{B}\cap\hat{\mathcal{H}}} w_{i} 
\label{unbiasedB}
\end{equation*} 
Events which are predicted by the classifier to be of the positive signal class but actually belong to the negative class, a false positive.}
\item{Selected Signal events : 
\begin{equation*} 
\hat{s}=\sum_{i \in \mathcal{S}\cap\hat{\mathcal{H}}} w_{i} 
\label{unbiasedS}
\end{equation*} 
Events which are predicted by the classifier to be of the positive signal class and do belong to the positive signal class, a true positive.}
\end{itemize}

The AMS function defined in \ref{ams} is computed on these quantities i.e. the expected number of signal and background events in the selection region. One way of describing the selection region is a region of the feature space where an excess of signal events is expected over background. Hence, a binary classifier for this task can be viewed as a tool for identifying signal-rich regions in the feature space. 

The occurrence of background events follows a Poisson process (in any part of the feature space, even in the selection region). Over a given time period during which events are recorded, the number of background events ending up in the selection region is $\mu_{b}$ and the variance is also $\mu_{b}$ (the mean and variance of a Poisson random variable are identical). The normalized statistic, 

\begin{equation} \hat{t} = (n-\mu_{b})/\sqrt{\mu_{b}} \sim N(0,1) 
\label{normal}
\end{equation} 

(where $n$ is the total number of events in the selection region) serves as a test statistic for detection of signal events. A fluctuation is considered sufficiently large to claim a discovery of the signal process if it exceeds $5\sigma$, i.e. if $\hat{t} > 5$ ($\sigma = 1$ for the normalized test statistic). A $5\sigma$ significance corresponds to a $p-$value of $3$ x $10^{-7}$. The magnitude of the $p-$value can be interpreted as a probability of observing a test-statistic as extreme or even greater given that the null hypothesis of background only was true. 

All events in the selection region of a classifier are predicted positives, this simplifies the test statistic further, $n$ which is the total number of events in the selection region is essentially $\hat{s}+\hat{b}$, and $\mu_{b}$ which is the expected number of selected background events (false positives) can be approximated by its empirical counterpart, $\hat{b}$. Substituting this in \ref{normal} gives, 

\begin{equation}
(n-\mu_{b})/\sqrt{\mu_{b}} = ( \hat{s} + \hat{b} - \hat{b})/\sqrt{\hat{b}} = \hat{s}/\sqrt{\hat{b}}
\label{simple}
\end{equation}

This is the simplified AMS metric, essentially a ratio of the true positives to false positives calculated based on the events in the selection region of a classifier.

\begin{sidewaysfigure}
\centering
\includegraphics[scale=0.6]{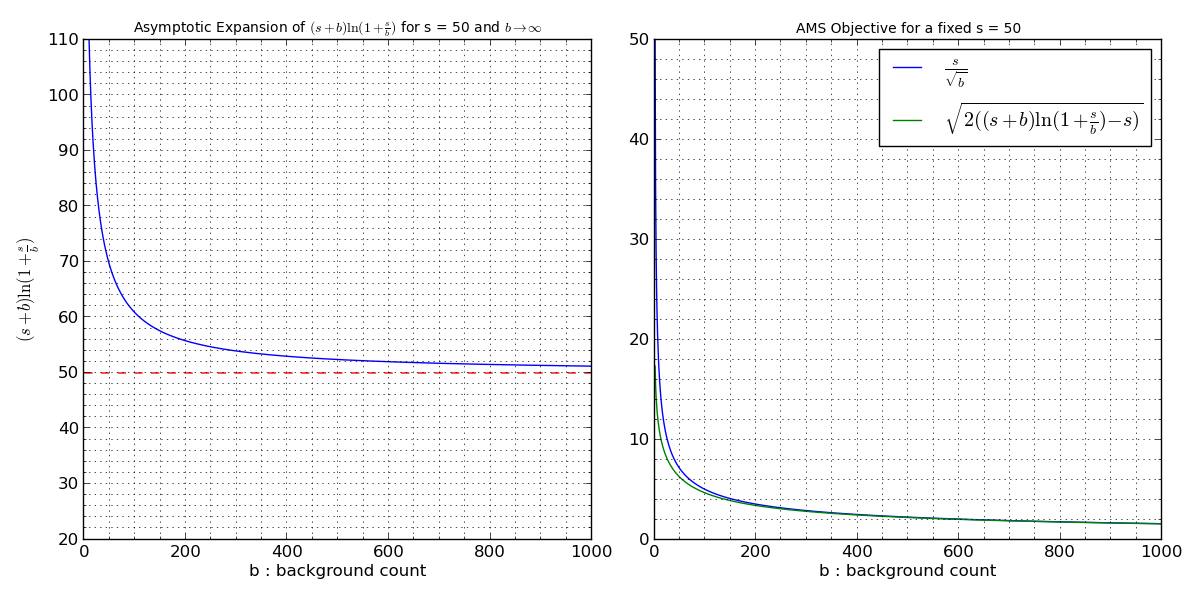}
\label{ams_theory}
\caption{Behaviour of the AMS in asymptote}
\end{sidewaysfigure}

It is worthwhile to note that when $s \ll b$  eq. \ref{simple} is equivalent to eq. \ref{ams} in its asymptotic expansion. Fig. \ref{ams_theory} depicts this for a fixed signal count of s=50.

\section{On p-values}
\label{pvalues}

\subsection{p-values: Interpretation}

$\pvalue$s were theorized by statistician R. Fisher and formalized in his book \textit{Statistical Methods for Research Workers}, published in 1925. They play a central role in hypothesis testing where one is seeking to accept or reject usually the null hypothesis and a specified mathematical model. The model is used to come up with instances or sample observations which are then numerically summarized in a single scalar value, called the \textit{test}-\textit{statistic}. The mathematical form of the test statistic is usually tied to the needs of the experiment. It is constructed so as to quantify from observed data, patterns that would distinguish null from alternative hypothesis. 

The $\pvalue$ is a function of the test statistic and measures the probability of observing values (of the test statistic) at least as extreme as the ones observed given that the null hypothesis is true. In any experiment, it is always possible that the observed value resulted due to a sampling error. The $\pvalue$ measures the probability of the observed effect being an artefact of the sampling process. Hence, lower $\pvalue$s are associated with more significant outcomes. 

The interpretation of a $\pvalue$ is done with the help of a significance threshold, typically denoted using the symbol $\alpha$. A significance threshold is chosen independently and for most experiments in social sciences a choice of 0.05 or 0.01 is rendered as good scientific practice. If the $\pvalue$ falls below the significance threshold, say $p < 0.05$ the results are deemed to be statistically significant under a 95$\%$ confidence interval (1-$\alpha$). Statistical significance is a necessary condition to reject the null hypothesis. The significance threshold captures the probability of rejecting the null hypothesis given that it is true, this is called committing a type I error (subtly different to the $\pvalue$). Hence, the lower the significance threshold $\alpha$, the more stringent the conditions for rejecting the null hypothesis. The initial choice of 0.05 as the cut-off level for significance was first proposed by R. Fisher and has persisted as the most popular initial choice for experiments to date. In particle physics, the threshold for ``evidence" of a particle is set at $p=0.003$ and the standard for discovery is $p=3 \times 10^{-7}$. 

It is important to clarify how $\pvalue$s relate to significance thresholds described in terms of ``sigmas" i.e. 3$\sigma$ or 5$\sigma$. In particle physics, a convention is followed to report the significance in units of standard deviations from the mean or sigmas which is equivalent to a specific $\pvalue$. A $\pvalue$ corresponding to 3$\sigma$ denotes the probability of sampling a value 3 standard deviations away from the mean in a gaussian distribution. 

\begin{figure}
\begin{center}
\includegraphics[scale=0.7]{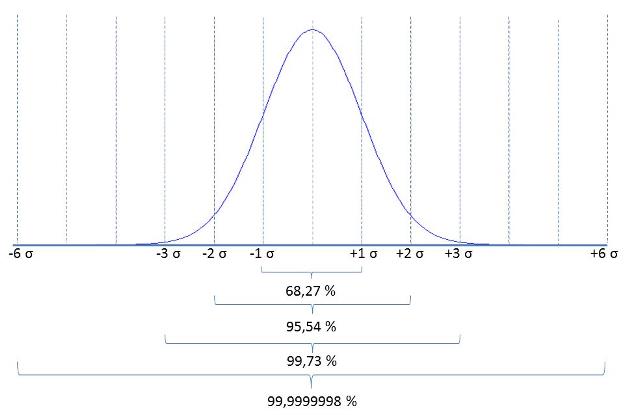}
\caption{Gaussian distribution}
\label{sigma}
\end{center}
\end{figure}

To give an idea of the rarity of picking such a sample, see fig.\ref{sigma} which shows that 99.73$\%$ of the observations are within $3$ standard deviations of the mean. 

As stated earlier, the $\pvalue$ is a function of a test statistic which is a scalar computed using all the observations recorded in an experiment. The calculation of a $\pvalue$ involves knowing the sampling distribution of the test statistic either exactly or approximately, in many cases the test statistic can be approximated by the gaussian distribution for large sample sizes due to the central limit theorem. For a random variable $X$, an experiment to test if the population mean $\mu$ was equal to a value say $\delta$ would have a test statistic of the form, 

\begin{equation}
Z = \frac{\bar{x} - \delta}{\sigma} 
\end{equation} 

where $\delta$ is the value to be tested against, $\sigma$ is the population standard deviation and $\bar{x}$ is the sample mean. The test statistic, called the Z-score in this case has a standard normal distribution and the $\pvalue$ is $(1 - \Phi(X < Z))$ where $\Phi$ is the cumulative probability distribution function of a standard gaussian variable. Therefore, $Z = \Phi^{-1}(1-p)$ and measures significance in terms of number of sigmas from the mean. 

\begin{figure}
\includegraphics[scale=0.6]{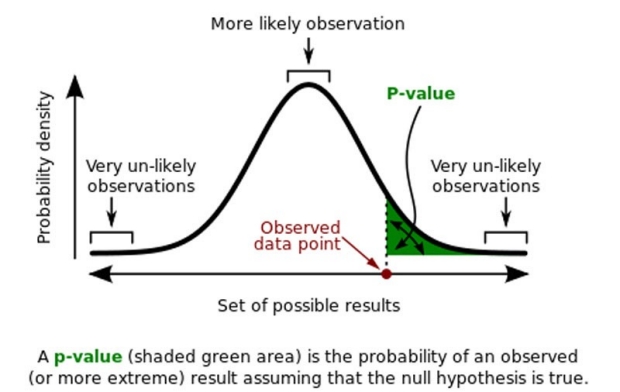}
\caption{p-value depiction}
\end{figure}

In the case of the Higgs discovery, a $5\sigma$ event corresponds to a $\pvalue$ of at least as low as $1 - \Phi(5) \approx 3 \times 10^{-7}$.
 
\subsection{Criticisms}

The role of using $\pvalue$s in the process of scientific discovery has been fraught with controversy and problems with interpretation.  The essence of the argument has been that a single metric like the $\pvalue$ does not quantify the credibility of a conclusion made on the basis of a hypothesis test. It is a statistical tool that can be used to indicate if a certain null hypothesis deserves further scrutiny. As stated in \cite{goodman}, refutation of a null hypothesis based on the $\pvalue$ crossing a significance threshold (effectively called "bright-line" thinking) to the exclusion of other factors that intrinsically impact the experiment like design, methodology and external evidence is a serious fallacy of interpretation.  

It is dangerous to use the $\pvalue$ crossing a significance threshold as a marker for truth. \cite{goodman} also states how R. Fisher who revolutionized inference in the context of frequentist statistics and formulated the original idea of the $\pvalue$ did not intend it to be used as a make or break metric. A single demonstration of a $\pvalue$ crossing a threshold is not enough to make a scientific claim unless repeated experiments under identical settings "rarely" failed to achieve that threshold. The important factor here is repeatability. Using it in any other way can lead to claims that are either false or largely overstated. 

There is hardly any statistical literature that explains why significance thresholds vary among disciplines. Apart from the stringency criterion which sets a high benchmark for the significance threshold in sciences like physics to be $p \leq 3 \times 10^{-7}$ (5$\sigma$)and in genomics to be $p \leq 10^{-8}$ (5.6$\sigma$) there isn't any analyses explaining how precisely these numbers came about. This calls for some caution in applying them as hard thresholds, effectively, there isn't a good reason to dismiss a $4.9\sigma$ claim just because it doesn't meet the cut-off.

\chapter{Performance Metrics: A discussion}
\label{performance}

Performance measures to be optimized through learning are as diverse as the number of learning methods. In supervised binary classification, the generalization error defined as the misclassification rate on unseen or new data is considered to be the benchmark metric over which classifiers are optimized. 
In the sections below we discuss the relationship of each performance measure with the AMS metric and assess its relevance for the classification task at hand. 

Herein, we refer to the background class (N) as belonging to the majority class with a negative label of -1 and the signal class (P) as belonging to the minority class with a positive label of +1. 

\section{Importance Weights}

All the metrics described in the sections below are computed using importance weights $w_{i}$ rather than the count. Similar to the concept of a selection region $\mathcal{\hat{H}}$ introduced in eq. \ref{index_select} (section \ref{math}, chapter \ref{formal}) it is useful to introduce the notion of a rejection region that contains all events that fall outside of the selection region. It can be defined as, 

\begin{equation}
\mathcal{H}^{c} = \{\mathbf{x} : h(\mathbf{x}) = b\} = \{\mathbf{x} : f(\mathbf{x}) < \theta_{k}\}
\label{rejection}
\end{equation}

The index set of points that belong to the rejection region is given by, 

\begin{equation}
\mathcal{\hat{H}}^{c} = \{i : \mathbf{x}_{i} \in \mathcal{H}^{c}\} = \{i : h(\mathbf{x}) = b\} 
\end{equation}

Further, $\mathcal{H} \cap \mathcal{H}^{c} = \emptyset$ and $|\mathcal{H} \cup \mathcal{H}^{c}| = n$ where $n$ is the total  number of events in the dataset. Each event after classification is either in the selection region or the rejection region. 

Using the definitions of selection region and rejection region it is easy to enunciate the most commonly used terms in a binary classification context using importance weights:

\begin{table}[H]
\begin{center}
\begin{tabular}{c|c|c|c|c}
& & \multicolumn{2}{c}{Predicted Label}\\
\hline
& & -1 (b) & +1 (s) \\
\hline
\multirow{3}{*}{True Label} & -1 (b) & TN: $\sum_{i \in \mathcal{B} \cap \hat{\mathcal{H}^{c}}} w_{i}$  & FP($\hat{b}$): $\sum_{i \in \mathcal{B} \cap \hat{\mathcal{H}}} w_{i}$  & TN + FP = N\\ 
& +1 (s) & FN: $\sum_{i \in \mathcal{S} \cap \hat{\mathcal{H}^{c}}} w_{i}$ & TP($\hat{s}$): $\sum_{i \in \mathcal{S} \cap \hat{\mathcal{H}}} w_{i}$ & FN + TP = P\\ 
\hline 
\end{tabular}
\label{confusion_weights}
\caption{Classification terminology using weights}
\end{center}
\end{table}

The values falling in the second column (TP and FP) are directly used to determine the AMS metric $\dfrac{\hat{s}}{\sqrt{\hat{b}}}$.

\section{Accuracy: Training and Test error}

The training error is the misclassification rate of training samples and test error is the misclassification rate when a trained classifier is applied to unseen data points. Accuracy is (1 - misclassification rate) expressed as a $\%$.

As stated earlier, the error rates are a weak indicator of performance in the presence of imbalanced prior class distributions. A class with 99$\%$ samples of the majority class will achieve a 99$\%$ accuracy rate with a classifier that blindly assigns all samples to the majority class. 

In the Higgs dataset used for training and testing approximately 70$\%$ of the data points belong to the majority background class hence a classifier which achieves a 70$\%$ accuracy rate on test data is not a useful indicator of performance unless the measure is scrutinized further. 

\section{Recall and Precision}

The recall metric also called \textit{sensitivity} is computed as the fraction of the data points of the positive signal class correctly predicted as positive. Using the terminology described in table \ref{confusion_weights}, it is the fraction $\dfrac{\text{TP}}{\text{P}}$.  

The precision metric, is computed as the fraction of the predicted signals which are actually signals. It is the fraction, $\dfrac{\text{TP}}{\text{TP}+\text{FP}}$.

Since the AMS metric constitutes true positives (TP) and false positives (FP), the precision metric is a good indicator of AMS performance. 

\section{Balanced Classification error}

We have established that the overall classification accuracy in terms of count of events is a weak indicator of the strength of a classifier in the presence of unbalanced classes. A metric that effectively captures the fluctuations in the AMS must incorporate the importance weights $w_{i}$. One that is proposed by ATLAS physicists is the \textit{balanced classification error}. It is defined as, 

\begin{equation} 
R(f) = \sum_{i=1}^{n}w_{i}'\mathbb{I}\{y_{i}^{pred} \neq y_{i}^{true}\}. 
\label{bce}
\end{equation}

$\mathbb{I}$ is the indicator function. The weights $w_{i}'$ are normalized in both the signal and background classes to $N_{b}' = N_{s}' = 0.5$, that is, 

\begin{equation}
w_{i}' = w_{i}  \hspace{2mm} \times 
\begin{cases}
\dfrac{1}{2N_{s}} & \textrm{ if } i \in \mathcal{S} \\
\dfrac{1}{2N_{b}} & \textrm{ if } i \in \mathcal{B} 
\end{cases}
\label{neutralize}
\end{equation}

$N_s$ and $N_{b}$ denote the expected number of background and signal events as described in eq. \ref{unbiased}. 

It is important to neutralize the weights in this manner to compute the classification error in order to penalize misclassified signals as severely as misclassified background events. The original weights $w_{i}$ for signal events are on average 300 times smaller than those for background events. The balanced weights are generated only for the purposes of calculating the balanced classification error metric $R(f)$, the AMS is \textbf{always} computed on the unbalanced weights as in eq. \ref{weights}.

\begin{figure}
\includegraphics[width=\textwidth]{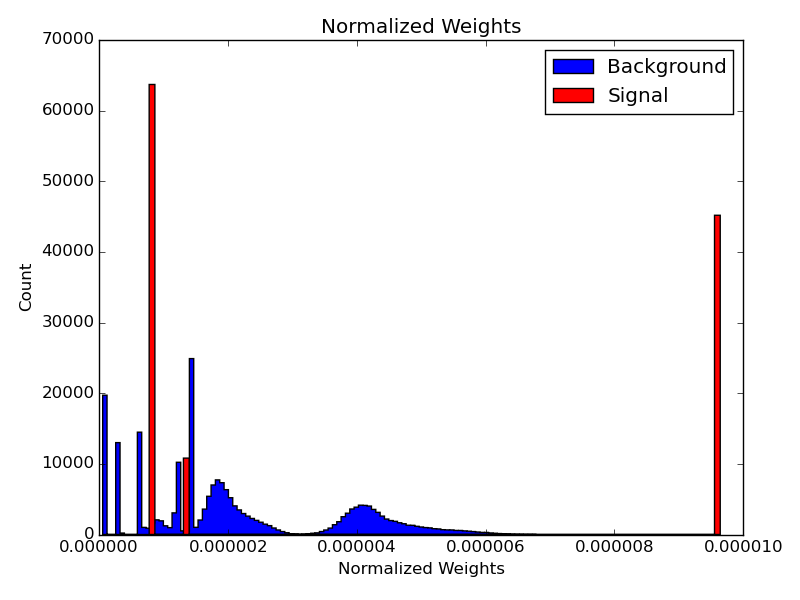}
\caption{Balanced weights for the signal and background classes.}
\label{bweights}
\end{figure}

Fig. \ref{bweights} shows the distribution of weights of the signal and background class after re-balancing them as per eq. \ref{neutralize}.

The magnitude of the balanced classification error of a classifier is a good indicator of AMS performance. Fine-tuning the parameters of a classifier to minimize this metric is an overwhelmingly popular approach. 

\section{Receiver Operator Characteristic (ROC)}

\begin{figure}
\begin{center}
\includegraphics[scale=3]{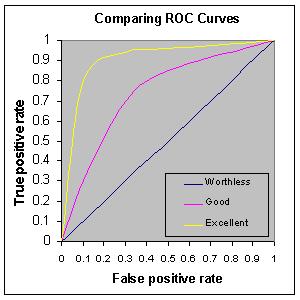}
\caption{ROC curves for classifiers with different levels of prediction accuracy}
\end{center}
\label{roc}
\end{figure}

It is interesting to analyse the relationship between the simplified AMS metric $\hat{s}/\sqrt{\hat{b}}$ and the \textit{Receiver Operating Characteristic} (ROC) curve \footnote{The rather unusual name ROC emerged during World War II for the analysis of radar images. Radar operators had to decide whether a blip on the screen was an enemy target, friendly ship or just noise. Signal detection theory measures the ability of radar receiver operators to make these import distinctions. Their ability to do so was called \textit{Receiver Operating Characteristics}.}, they are closely related but not correlated. The ROC curve illustrates the performance of a binary classifier by depicting the true positive rate (TPR = TP/P), against the false positive rate (FPR = FP/N). A fixed threshold $\theta_{k}$ gives a single TPR and FPR (a single point on the curve), the curve is generated by computing the TPR and FPR for different values of $\theta_{k}$. Fig. \ref{roc} is an example of ROC curves for 3 different classifiers. 

The $45^\circ$ line denotes a random classifier, which at no threshold gives a higher TPR than FPR. A ROC curve that lies above the $45^\circ$ denotes a classifier with higher than random classification accuracy of positive samples for all values of the threshold and encloses a larger area under the curve. A perfect classifier has a TPR = 1 and FPR = 0 denoting perfect accuracy. The closer the ROC curve for a classifier is to the upper left corner of the graph (1,0) the better the classifier. The value on the $x$-axis, the FPR is also expressed as (1 - \textit{specificity}), where  specificity is the true negative rate (TN/N). Optimizing a classifier for the ROC curve pulls the curve towards the upper left corner of the graph to give higher true positive rates for each false positive rate.   

It seems as though then the threshold $\theta_{k}$ which corresponds to the upper left most point on the ROC curve should correspondingly maximize the AMS, however that is not the case and it is not immediately apparent as to why. 

I offer two explanations for this.

\begin{enumerate}

\item By ROC standards when choosing between two classifiers, the classifier that generates a higher ROC curve and encloses a greater area underneath it is the more optimal one, the area under the ROC curve is shortened as ROC$\_$AUC. The ROC$\_$AUC integrates across all possible choices of the threshold $\theta_{k}$. The threshold that corresponds to the upper left most point on the curve is chosen as the optimal threshold $\theta_{k}$. 
By AMS standards, we are concerned with the TPR and FPR values at a single optimal choice of threshold $\theta_{k}$. Effectively, this is a single point on the ROC curve. The shape or height of the ROC curve does not matter. We can achieve the same optimal AMS value on two very different ROC curves. 

\item Further, there is no guarantee that the AMS is maximized at the upper left most point of the ROC curve. The upper left most point on the ROC curve is the point which maximizes the ratio TPR/FPR. This ratio is the slope of the tangent to the ROC curve at a single point. The next section describes this ratio and contrasts it with the behaviour of the AMS.   

\end{enumerate}

In summary, optimizing for ROC is not equivalent to optimizing for AMS. 

\section{Likelihood Ratio}

The slope of the tangent line to the ROC curve at a fixed threshold $\theta_{k}$ is the ratio $\dfrac{\textrm{TPR}}{\textrm{FPR}}$. This is also called the positive likelihood ratio $(\textrm{LR}+)$, 

\begin{equation} 
\textrm{LR}+ = \frac{\textrm{TPR}}{\textrm{FPR}} = \frac{sensitivity}{(1 - specificity)} 
\end{equation}

It is easy to see that this ratio is maximised at the extreme upper-left hand corner of the ROC curve, this is the point that gives the best trade-off between  the true positive rate and false positive rate. It is a reasonable approach in classification to maximise the LR+ of a classifier in order to improve its overall classification accuracy.

Recall that the AMS metric $(\hat{s}/\sqrt{\hat{b}})$ is essentially the ratio of true positives to false positives in a selection region $\mathcal{H}$ specified by a cut-off threshold $\theta_{k}$. Maximizing the AMS is tantamount to maximizing the true positives and minimizing the false positives in the selection region. This seems very close to the idea of maximizing the positive likelihood ratio $(LR+)$. However, there are some fundamental differences.

\begin{enumerate}
\item The likelihood ratio uses true positive and false positive rates while the AMS uses unnormalized sums.

\begin{equation}
\textrm{ LR+: } \dfrac{\hat{s}/N_{s}}{\hat{b}/N_{b}} = \frac{\sum_{i \in \mathcal{S}\cap\hat{\mathcal{H}}} w_{i}/\sum_{i \in \mathcal{S}}w_{i}}{\sum_{i \in \mathcal{B}\cap\hat{\mathcal{H}} w_{i}}/\sum_{i \in \mathcal{B}}w_{i}}
\end{equation}

\begin{displaymath}
\textrm{ AMS: }\frac{\hat{s}}{\sqrt{\hat{b}}} = \frac{\sum_{i \in \mathcal{S}\cap\hat{\mathcal{H}}} w_{i}}{\sqrt{\sum_{i \in \mathcal{B}\cap\hat{\mathcal{H}}} w_{i}}}
\end{displaymath}

This distorts the correlation between the likelihood ratio and AMS metric. It is possible to achieve a higher AMS metric at a point on the ROC curve where the LR+ ratio is not maximized.

\item{The AMS ignores all samples that lie in the rejection region like false negatives (signals predicted to be background), however, LR+ is sensitive to it,
\begin{center}
TPR = TP/P = TP/(TP + FN)
\end{center}} 
The AMS using these metrics is simply, TP/$\sqrt{\textrm{FP}}$.

\end{enumerate}

\section{AMS (\texorpdfstring{$\sigma$}{s})}

\label{metrics}

A classifier is trained to minimize the balanced classification error as in eq. \ref{bce}. The AMS is then optimized with respect to a threshold $\theta_{k}$ in the classifier that generates a selection region $\mathcal{H} = \{\mathbf{x}: f(\mathbf{x}) > \theta_{k}\}$ with predicted signals. This is tantamount to classifying according to the rule,  $\sign\left(f\left(\mathbf{x}\right) - \theta_{k}\right)$. Prior experiments at ATLAS suggest that the AMS is maximized at a threshold $\theta_{k}$ yielding a selection region of the top $15\%$ of the events ranked by score. This implies selecting $\theta_{k}$ as the 85th percentile value of the score $f(\mathbf{x})$. Direct optimization of the AMS metric is infeasible as it is fully determined by the small number of events in the selection region $\mathcal{H}$, this makes it noisy and ill-conditioned since a small perturbation in the classifier can lead to a different composition of events in the selection region and a different value of the AMS. The suggested approach is a two step process. 

\begin{enumerate}
\item  Fine tune the classifier performance treating the balanced classification error as a loss function. 
\item Fine tune the choice of threshold $\theta_{k}$ for the classifier optimized in Step 1 through optimization of the AMS.   
\end{enumerate}

From a machine learning point of view the Higgs dataset represents two fundamental challenges described in the sections below.

\section{Class imbalance \texorpdfstring{\&}{} Class overlap}

\begin{figure}[H]
\centering
\includegraphics[scale=0.5]{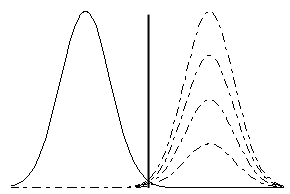}
\caption{Well separated classes}
\label{less}
\end{figure}

\begin{figure}[H]
\centering
\includegraphics[scale=0.5]{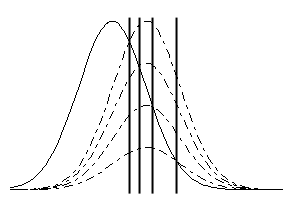}
\caption{Overlaid classes}
\label{more}
\end{figure}

Studies show that there are at least two main sources of difficulty in classification problems - one of them is class overlap and the other is class imbalance. Class imbalance refers to the under-representation of one or more class labels in the training and test data .i.e., a difference in class prior probabilities. Class overlap exacerbates the problem of class imbalance. In order to illustrate the conjecture consider the case of a single attribute problem with two classes. The probability distribution of the attribute values of the two classes are given by the gaussian distributions with the same variance but different means. In figure \ref{more} below the mean of the class represented by the dashed line (positive class) is one standard deviation away from the mean of the class represented by the solid line (negative class). The vertical lines denotes the optimal split. It is easy to see that the influence of changing prior probabilities of the positive class on the optimal split is much greater in the highly overlaid instances, in fig. \ref{more} than in fig. \ref{less} where the instances are well separated. In fig. \ref{less} where the mean of the positive class represented by the dashed line is four standard deviations away from the negative class, the optimal split derived by a simple classifier is inelastic to the changing priors \cite{overlap}.

\section{Bias-Variance trade-off}

In machine learning parlance the bias-variance trade-off refers to the problem of balancing model accuracy and model generalization ability, it is usually impossible to improve both simultaneously. A highly accurate model that captures the regularities of the training data well is a low-bias model. A low-bias model comes at the cost of high model complexity which can lead to the problem of over-fitting. An over-fitted model is sensitive to small fluctuations in the training data set. Even a slight perturbation of the training data leads to a substantially different model structure. Hence, low bias usually occurs with high variance. 

On the flip side, a simple model which is more deterministic in its output is a high-bias low-variance model. It is too simple to capture the hidden relationships between the training data and output leading to high-bias (low accuracy) and under-fitting but has low variance. An under fitted model is stable to small perturbations in the training data. Hence, high bias usually occurs with low variance.

\begin{figure}[ht]
\begin{center}
\includegraphics[scale=0.5]{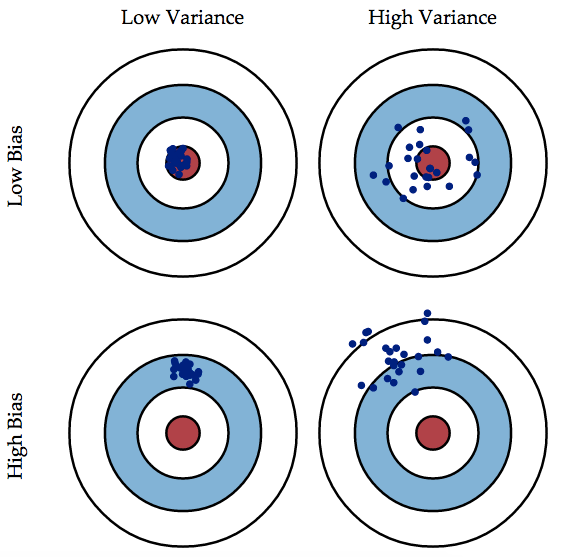}
\caption[Bias-Variance Depiction]{This figure illustrates the different combinations of bias and variance. The red  center represents the target that the points need to fit to. The concentric circles represent different levels of diffusion. Low bias-low variance models present a theoretical standard - this is where models want to be. High-bias high variance models represent the attributes of an ill-defined model. Most models fall into the other two categories of being either low-bias high-variance or high-bias low-variance.}
\end{center}
\end{figure}

The error or misclassification rate on the training data is usually used to measure bias of a model and the variance of the output predictions of a model capture the variance of a model. The negative relationship between bias and variance of a model make it difficult to moderate both measures simultaneously. 

Robust models usually lie somewhere in the middle of the bias-variance spectrum with an acceptable amount of bias and variance. The goal of parameter optimization is usually to find this point on the bias-variance trade-off curve.

\begin{figure}[ht]
\includegraphics[width=\textwidth]{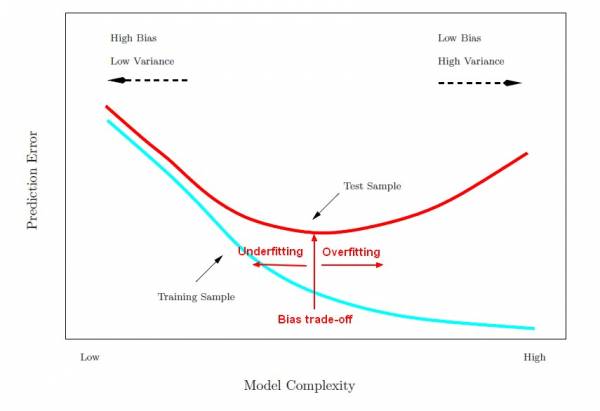}
\caption[Bias-Variance vs. Model Complexity]{The red curve depicts prediction/generalization error, the blue curve depicts training error. The diagram depicts that it is possible to achieve arbitrarily accurate models in fitting to a training dataset, such models usually define complex rules to capture the hidden relationships in training data. However, the effectiveness of such models on unseen data is poor as they are too over-fitted to the training set. There is a cross-over point where the error on test data starts to degrade. Good models position themselves at the point when this starts to happen.}
\label{complexity}
\end{figure}

\chapter{On meta-algorithms: Bagging \texorpdfstring{$\&$}{} Boosting}
\label{ad}

The algorithm developed in this thesis provides a way to combine two popular theories in machine learning - boosting and bagging. Both techniques provide a framework for ensemble learning and are meta-algorithms rather than learners. Rather than conducting the learning themselves, boosting and bagging are techniques that can be applied to rudimentary learners to achieve better performance. A learner can be defined as an algorithm with an underlying mathematical formulation to map input to output. A decision tree in the context of classification is an example of a learner. There are different ways to define meta-learning, I choose to define it in the following way. A meta-learner is a heuristic device that provides an architecture to learn from the output of learners (one or multiple). The algorithm proposed in section \ref{comb} of this chapter fits the definition of a meta learning system. In the next few sections I provide a short account of the primitive binary tree learner and a more detailed account of the boosting and bagging principles. The mathematical formulation of primitive tree learning is beyond the scope of this study.

\section{Tree learning}
\label{DT}

Decision trees (DT) are predictive and non-parametric models which use supervised learning for the task of classification. Given a set of training points $\{\textbf{x}_{i},y_{i}\}$ where $\textbf{x}_{i} \in \mathbb{R}^d$ is an input feature vector and $y_{i}$ is a categorical class variable the DT learns simple rules inferred from the input features with the goal of mapping them to their correct class labels.Typically, a decision tree has a top-down flow-chart like structure. 

\begin{figure*}
\hspace{-10mm}
\includegraphics[scale=0.4]{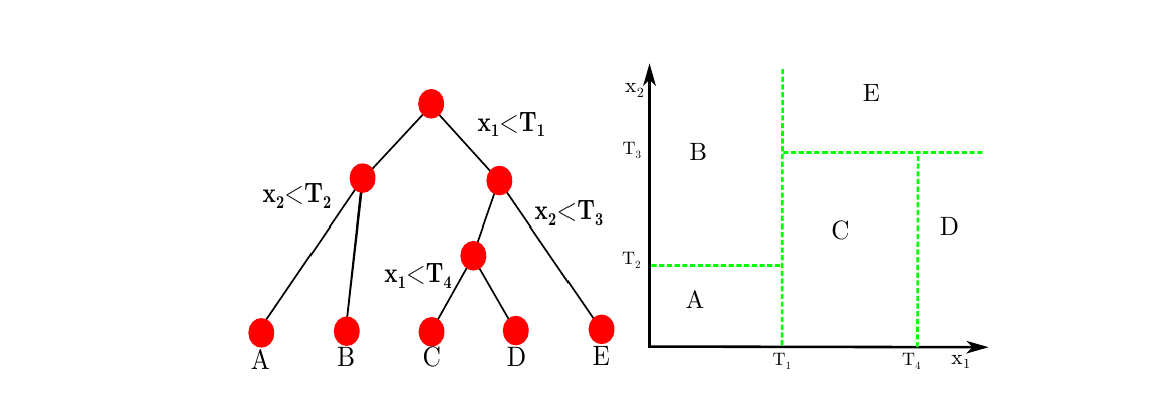}
\caption{Tree learner}
\label{flow_tree}
\end{figure*}

At the top of the tree is a single \textit{source} node which at the beginning of the learning process has all the training data points. A tree starts learning by splitting the source node on the basis of a value test on a chosen input feature attribute. The value test essentially applies a threshold on the value of the chosen feature attribute and   partitions the training set. The partitions are expressed in the form of branches that emerge from the source node. This process is repeated in a recursive manner on each of the derived subsets. Fig \ref{flow_tree} illustrates the procedure of recursive partitioning. The choice of feature attribute to split on and the threshold for the value test are tied to user specified splitting criterion. The figure on the right shows the decision boundary that emerges as a result of applying univariate splits to the data. New data are classified on the basis of which rectangular region they fall into. In a high-dimensional feature space a decision tree gives hyper-rectangular regions. The tree in the figure has a depth of 4 and the nodes [A,$\ldots$,E] represent leaves, these might not be pure (i.e. contain samples of one class). The leaves are assigned a class based on the majority of samples that end up in that leaf.
 
The recursive phase of the tree continues until a stopping criterion is triggered. Some of the most common stopping rules are, 

\begin{enumerate}
\item{All of the training samples in a partition belong to the same class -- in this case, the node is converted to a leaf and recursion continues on the other branches.}
\item{User specified maximum depth has been reached.}
\item{A partition has less than the minimum number of samples required for a split, this parameter can be user specified, the default value is 2.}
\item{The best splitting criterion is not greater than a certain threshold.}
\end{enumerate} 

\section{Ensemble learning} 
\label{ensemble}

A single-pass classifier is a classifier that processes each event in the dataset once - the supervised training step is conducted by fitting a model to the whole dataset which is provided in the beginning with true labels. The decision tree introduced in the previous section is a classic example of a single pass classifier. This can be contrasted with ensemble classifiers that involve multiple iterations of training. In ensemble techniques training typically occurs through using a heterogeneous mix of classifiers or using copies of the same classifier initialized with different parameter settings. 

An example of an ensemble technique would be to train the dataset on a decision tree and a support vector machine and finally combine their predictions according to a certain weight to reach classification decisions. Ensembles can be built out of any diverse number of base learners. 

The ATLAS Higgs dataset is characterized by densely overlapping features (class overlap in the binary context was defined in chapter \ref{performance}). Popular published approaches that deal with the problem of class overlap suggest that the presence of highly overlaid classes significantly degrade the performance of single pass classifiers and point to the need for more sophisticated techniques like \textit{ensembles}. While ensemble techniques are known to be very effective in improving performance their main criticism is that techniques to combine different classifiers are not rooted in mathematical theory. The choice of base learners are guided by empiricism - characteristics of the data, bounds on model complexity, time and memory.

In the next sections we discuss two of the most popular themes in ensemble learning - boosting and bagging.  

\section{Boosting: concept \texorpdfstring{$\&$}{} description}

The aim of this section is to present the concept of boosting and describe one of the most fundamental expressions of the idea - the AdaBoost algorithm \cite{scapire} won the 2003 G$\ddot{o}$del prize. 

The success of boosting techniques is attributed to the way they tackle the problem of class overlap. Boosting works by iteratively training `weak' learners to build a master learner which is significantly better at the task. The decision rule for the master learner is a weighted combination of the weak learners outcomes and the weights are usually a function of the weak learners accuracy. A weak learner is a classifier that can classify samples better than random guessing.

Boosting ensures that samples in the most ambiguous regions of the feature space which are repeatedly misclassified are encountered in multiple stages of the training process, thereby allowing the classifier to focus on the hard to classify events. This explains how boosting achieves low bias. At each iteration it improves accuracy by classifying few more points than in the previous stage correctly without disrupting the correct classifications from previous stages. 

While boosting is considered to be one of the best off-the-shelf classifiers in the world it is prone to overfitting the training data. This makes it a high variance classifier. Boosting with several stages of tree learning can rapidly overfit as trees are low bias high variance learners. To control over-fitting while using boosting is one of the main challenges in using this technique. Boosted decision trees (BDTs) use decision trees (or decision stumps - a one-level decision tree) as base learners within the context of boosting and have been the benchmark algorithm for multivariate analysis and classification problems for particle identification at CERN. 

\subsection{AdaBoost}

AdaBoost is a meta-algorithm which uses the principle of boosting. It has a cascade architecture that trains the data in multiples stages where each stage uses information from the output of the previous stage. AdaBoost initiates the first stage by assigning a uniform weight to each training sample. At the end of the first stage it increases the weight on the misclassified samples in a weight update step. This process is continued until the misclassification rate saturates. 

Consider a binary classification problem with input vectors $\mathbf{D} = \{(\mathbf{x}_{1},y_{1},w_{1}) \ldots (\mathbf{x}_{n},y_{n},w_{n})\}$ with binary class labels $y_{i} \in \{-1,+1\}, \forall i=1 \ldots n$. 

At the start of the algorithm, the training data weights $\{w_{i}\}$ are initialized to $w_{i}^{(1)} = 1/n, \forall i=1 \ldots n$. A base classifier $h_{1}(\mathbf{x}): \mathbf{D} \rightarrow \{-1,+1\} $ that misclassifies the least number of training samples is chosen. Formally, $h_{1}(\mathbf{x})$ minimizes the weighted error function given by, 

\begin{equation}
R(h_{1}) = \sum_{i=1}^{N}w_{i}\mathbf{I}(h_{1}(\mathbf{x}) \neq y_{i})
\end{equation}
where $\mathbf{I}$ is the indicator function.  
 
After the first round of classification,  the coefficient $\alpha_{1}$ is computed that indicates the confidence in the classifier. It is chosen to minimize an exponential error metric given by,

\begin{align*}
E &= \sum_{i=1}^{N}e^{y_{i}\alpha_{1}h_{1}(\mathbf{x}_{i})}\\
&= \sum_{y_{i}\neq h_{1}(\mathbf{x}_{i})}e^{\alpha_{1}} + \sum_{y_{i} = h_{1}(\mathbf{x}_{i})}e^{-\alpha_{1}}
\end{align*}

\begin{align*}
\frac{dE}{d\alpha_{1}} &=  \sum_{y_{i}\neq h_{1}(\mathbf{x}_{i})}e^{\alpha_{1}} - \sum_{y_{i} = h_{1}(\mathbf{x}_{i})}e^{-\alpha_{1}} \\
&\Rightarrow \sum_{y_{i}\neq h_{1}(\mathbf{x}_{i})}e^{\alpha_{1}} = \sum_{y_{i} = h_{1}(\mathbf{x}_{i})}e^{-\alpha_{i}}\\
&\Rightarrow (N - N_{c})e^{\alpha_{1}} = N_{c}\frac{1}{e^{\alpha_{1}}}   \\
&\textrm{where $N_{c}$ is number of correctly classified samples by $h_{1}$}\\
&\Rightarrow e^{2\alpha_{1}} = \frac{N_{c}}{N - N_{c}}\\
&\Rightarrow \alpha_{1} = \frac{1}{2}\ln\bigg(\frac{N_{c}}{N-N_{c}}\bigg)
\end{align*}

Denoting $\epsilon_{1} = \dfrac{N - N_{c}}{N}$ as the error rate for $h_{1}$,

\begin{align}
\alpha_{1} = \frac{1}{2}\ln\bigg(\frac{1-\epsilon_{1}}{\epsilon_{1}}\bigg)
\label{alphas}
\end{align}

The weight update equation at each stage is given by, 

\begin{align*}
w_{i}^{(m+1)} = w_{i}^{(m)}e^{\alpha_{m}\mathbf{I}(h_{m}(\mathbf{x}_{i}) \neq y_{i})}
\end{align*}

The master learner $M_{h}(\mathbf{x})$ for a $M$ stage classifier is given by,

\begin{equation}
M_{h}(\mathbf{x}) = \sign\bigg( \sum_{m=1}^{M}\alpha_{m}h_{m}(\mathbf{x})\bigg)
\label{master}
\end{equation}
 
The intuitive idea behind AdaBoost is that by increasing the weights on misclassified samples successively, the classifier places greater emphasis on samples that are hard to classify. Traditional choices for weak learners in AdaBoost are logistic regression, decision stumps, decision trees and naive bayes classifiers. 

\begin{figure}[ht]
\includegraphics[width=\textwidth]{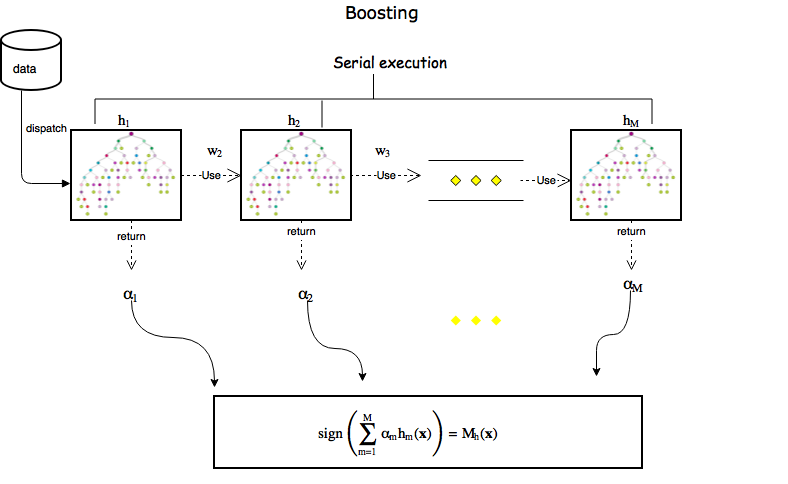}
\label{boosting}
\caption{Data Flow in boosting with single trees. In each iteration the weights $w_{i}$ are updated to focus more on the events misclassified in the previous stages. The $\alpha_{i}$s are the learning rates which are tied to performance of each base learner.}
\end{figure}

\section{Bagging: concept \texorpdfstring{$\&$}{} description}

Bagging derives from the words \textbf{b}ootstrap \textbf{agg}regat\textbf{ing} and was a technique popularised by Leo Brieman who first published the idea in 1996 \cite{bagging}. The bagging technique helps stabilise classifiers by training a single classifier on multiple bootstrap samples of the training data set. We end up with multiple versions of the same classifier which only differ in their training sets. A bootstrap sample is a result of drawing a uniform random sample \textbf{with replacement} from a dataset of size $n$. The size of the bootstrap usually is the same as the original dataset. Due to sampling with replacement a bootstrap sample most likely has repeated samples, this increases the predictive force of the base learner applied on those samples. Sampling sufficiently large number of times ensures that each element appears as duplicates in atleast one of the bootstrap samples.

Bagging can improve accuracy and is most useful when applied to unstable and high variance base learners, like decision trees. The instability of the base learner is the key requirement. 

Brieman notes:

``\textit{The vital element is the instability of the prediction method. If perturbing the learning set can cause significant changes in the predictor constructed, then bagging can improve accuracy.}"

Gains from bagging cannot be realized for classifiers which are stable and deterministic. Procedures like $k$-nearest neighbour are stable as they predict results based on a fixed spatial distribution of the training data. 
   
Given a training dataset $\mathbf{D} = \{(\mathbf{x}_{1},y_{1},w_{1}) \ldots (\mathbf{x}_{n},y_{n},w_{n})\}$ with binary class labels $y_{i} \in \{-1,+1\}$ and $w_{i} \in \mathbb{R}$  . Let $t: \mathbf{D} \rightarrow \{-1,+1\}$ be the decision tree learner that assigns an input vector $\mathbf{x}_{i}$ to a predicted label $\hat{y}_{i}$ and $\{\mathbf{D}^{(b)}\}$ be the sequence of bootstrap samples taken from $\mathbf{D}$. Each input $\mathbf{x}_{i}$ may appear in repeated bootstraps $\{\mathbf{D}^{(b)}\}$ or in none at all. The aggregated forest $t_{agg}$ is an expectation over individual trees $E(t)$. If $t$ predict a class label $\{-1,+1\}$, then $t_{agg}$ aggregates through majority voting. If $t$ predicts a probability then, $t_{agg}$ uses an arithmetic average of probability estimates from individual tree outputs $t(\mathbf{x}_{i})$. The number of bootstrap samples to use in bagging is usually established through computing the out-of-bag \footnote{OOB error refers to the prediction error of each sample $\mathbf{x_{i}}$ using trees which did not choose $\mathbf{x_{i}}$ in the bootstrap.} (OOB) error and stopping at the point at which the OOB saturates. A bootstrap sample $\{\mathbf{D}^{(b)}\}$ of size $n$ from $D$ will contain approximately 63.2$\%$ unique samples. 

The probability of an arbitrary sample $\mathbf{x}_{i}$ not being picked in $n$ draws is $(1-1/n)^{n}$. The result then follows from ($1 - \lim_{n \rightarrow \infty}(1 - 1/n)^n) =  (1 - 1/e) \approx 0.632$.

\begin{figure}
\centering
\includegraphics[width=\textwidth]{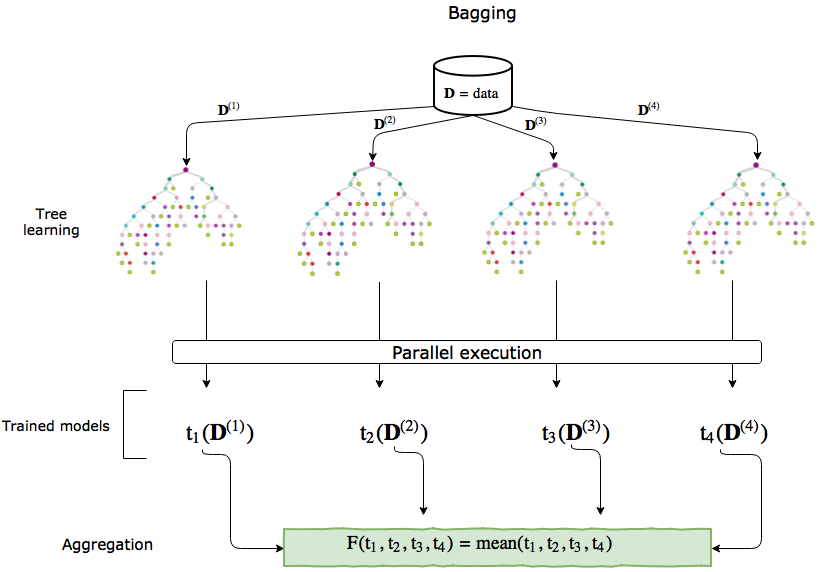}
\label{bagging}
\end{figure}

\subsection{Random Forest (RF)}

\gls{RF} is the trademark algorithm exemplifying bagging principle. A random forest as the name suggests is a collection of tree learners each trained on a bootstrap sample $\mathbf{D}^{(b)}$ of the training data $\mathbf{D}$. There are two sources of randomness in a random forest, the formation of bootstrap samples and the random selection of features considered at each node in the context of growing a single decision tree \cite{subspace}. The difference between the general bagging scheme for trees and a random forest is the second source of randomness also called $feature$ bagging. Many times, if there are $s$ features $\sqrt{s}$ (rounded down) features are randomly picked for consideration at each node.

The performance of a random forest depends on the strength of individual trees in the forest and the correlation between them. Perturbing the training set $\mathbf{D}$ by introducing duplicates through bootstraps is one way of decorrelating the trees however more different trees can be constructed if a random subset of features is selected at each split. In high dimensional data with a large number of features, this expands the universe of possible trees that can be constructed and introduces more tree dissimilarity.

Overall, the only disadvantage random forests have relative to a single pass decision tree is the loss of interpretability of the forest structure. A single decision tree has the nice property of being fully interpretable, a tree is essentially a `white-box' model which can be de-constructed  compared to models like artificial neural networks where the input to output flow cannot be easily unwound.

\begin{algorithm}[h]
\caption{Random Forest with $M$ trees}
\begin{algorithmic}[1]
\STATE \textbf{Training}:
\STATE \textbf{Input}: Training set $\mathbf{D}$, tree learner $t$ 
\STATE \textbf{Output}: Forest of trees $F(t_{1} \ldots t_{M})$ 
\item[]
\FORALL{$b$ = 1 \ldots $M$}
\STATE  \textbf{Construct} a bootstrap sample ${D^{(b)}}$ (sample uniformly with replacement from $\mathbf{D}$)
\STATE \textbf{Fit} tree learner $t_{b}$ on sample $D^{(b)}$
\ENDFOR
\RETURN $F(t_{1} \ldots t_{M})$  
\item[]
\STATE \textbf{Testing}:
\STATE \textbf{Input}: Test set $\mathbf{L} = \{\mathbf{x}_{i}\}_{i=1}^{n}$, forest of trees $F(t_{1} \ldots t_{M})$ 
\STATE \textbf{Output}: Predicted labels $\{\hat{y_{i}}\}_{i=1}^{n}$ 
\item[]
\FORALL {$\mathbf{x}_{i} \in \mathbf{L}$}
\STATE \textbf{Predict} $\hat{y}_{i} = F(\mathbf{x}_{i}) = F(t_{1}(x) \ldots t_{M}(x)) = E_{M}(t(\mathbf{x_{i}}))$
\ENDFOR
\RETURN $\{\hat{y_{i}}\}_{i=1}^{n}$
\end{algorithmic}
\label{rf}
\end{algorithm}

\section{Boosting vs. Bagging}

The meta techniques boosting and bagging specialize in two different aspects of the learning problem. The main focus in boosting is reducing bias and converging towards higher accuracy rates, it does so by narrowing focus on the hard to classify samples by repeatedly scaling up weights on these samples. While this gives powerful master learners, it is not resistant to the problem of over-fitting, which means that while they have the ability to fit very well to training data this needs to be balanced with their ability to generalize on unseen data. 

In bagging the main focus is variance reduction. A single unstable model is sensitive to noise in the training data and as a stand alone might fail to meet performance requirements however an average of several hundred unstable models which are uncorrelated is less sensitive to noise. A collection of trees trained on the same training dataset will give correlated results and the gains from averaging are subdued, however, a collection of trees each trained on a different training dataset gives de-correlated models. 

Boosting has no random elements, it goes from one stage of learning to the next and transfers an updated weight distribution which is fully determined by performance in the prior stage of learning. The weight distribution is adjusted to give misclassified samples more weight. The higher the weight the greater the influence of the sample on the learned model. Boosting is a closed form model, it only requires specification of the base learner and the number of stages of boosting. In contrast, bagging works because of the effectiveness of randomization. Randomization in creating bootstrap samples and feature subspace selection generate diversity in models and aggregating them helps overcome the instability of singular models. 

\section{Extremely Randomized Trees (ET)}

In the world of meta algorithms, bagging and boosting dominated the sphere for most of the nineties. There are several publications devoted to comparing the performance of each of the techniques on several classification datasets. \gls{ET} is one of the newer incarnations of the random forest method. The paper on ET was published in 2005 by  Geurts, Ernst and Wehenkel \cite{xrf}. The main principle is the introduction of randomization in the tree construction process. While random forests attempt to create diversified trees by using a bootstrap sample, ETs work like random forests but take the randomization one step further by choosing a random split point at each node rather than searching for the best split. This ensures the creation of strongly decorrelated trees. From the viewpoint of computational efficiency, ETs offer an advantage as they do not need to look for the most optimal split at each node. 

The random split generating algorithm from \cite{xrf} is summarized in \ref{randomsplits}. 

\begin{algorithm}[H]
\caption{Random Splitting algorithm}
\begin{algorithmic}[1]
\STATE \textbf{Input}: Training set $\mathbf{D}$
\STATE \textbf{Output}: A split $c_{j}$ which is a scalar value from a single feature vector of the training set $\mathbf{D}$  
\item[]
\STATE \textbf{Select} $K$ features ${a_{1}, \dots , a_{K}}$ at random from $\mathbf{D}$
\STATE \textbf{Select} $K$ splits $\{c_{1}, \dots , c_{K}\}$, one per feature for the $K$ features chosen in the previous step; each $c_{i}$ is selected at random from the range of the feature values $\forall i = 1, \dots K$  
\STATE Rank the splits $c_{i}$ by a criterion say $Q$ which gives a score $Q(\mathbf{D}, c_{i}) \in \mathbb{R}$ for each split. 
\STATE Select $c_{*} = max_{i=1 \dots K}Q(\mathbf{D}, c_{i})$  
\RETURN $c_{*}$
\end{algorithmic}
\label{randomsplits}
\end{algorithm}

A forest of extremely randomized trees trains each randomized tree on a bootstrap sample $\mathbf{D}^{(b)}$ and averages the output of randomized trees.

\begin{algorithm}[H]
\caption{Extremely Randomized $M$ trees}
\begin{algorithmic}[1]
\STATE \textbf{Training}:
\STATE \textbf{Input}: Training set $\mathbf{D}$, tree learner $t$ 
\STATE \textbf{Output}: Collection of extremely randomized trees $\{t_{1} \ldots t_{M}\}$
\item[]
\FORALL{$b$ = 1 \ldots $M$}
\STATE \textbf{Construct} a bootstrap sample ${D^{(b)}}$ (sample uniformly with replacement from $\mathbf{D}$)
\STATE \textbf{Fit} tree learner $t_{b}$ on sample $D^{(b)}$ using the Random Splitting Algorithm (see \ref{randomsplits})
\ENDFOR
\RETURN Collection of fitted trees $\{t_{1} \ldots t_{M}\}$  
\item[]
\STATE \textbf{Testing}:
\STATE \textbf{Input}: Test set $\mathbf{L} = \{\mathbf{x}_{i}\}_{i=1}^{n}$, collection of extremely random trees $\{t_{1} \ldots t_{M}\}$ 
\STATE \textbf{Output}: Predicted labels $\{\hat{y_{i}}\}_{i=1}^{n}$ 
\item[]
\FORALL {$\mathbf{x}_{i} \in \mathbf{L}$}
\STATE \textbf{Predict} $\hat{y}_{i} = agg(t_{1}(\mathbf{x_{i}}), \ldots, t_{M}(\mathbf{x_{i}}))$
\ENDFOR
\RETURN $\{\hat{y_{i}}\}_{i=1}^{n}$
\end{algorithmic}
\label{eforest}
\end{algorithm}

\section{Combining Bagging and Boosting} 
\label{comb}

There are several works which compare the effectiveness of bagging and boosting on several datasets. The consensus is that no technique is superior to the other and the effectiveness of the technique is tied to the characteristics of the problem at hand. For final testing on the Higgs dataset we propose an algorithm that is a variant of the boosted decision tree.  

The boosted ensemble works as a sequential learner but at each stage is supported by a bagged ensemble of primitive trees. There is very little to no evidence of models where bagging and boosting are harnessed together as such an ensemble is hard to optimize and not easily interpretable. We show that using extremely randomized trees as a base learner to boosting shows strong performance in the context presented by the classification task on the Higgs dataset. Simultaneously, we show that using a traditional random forest as a base learner to boosting does not prove to be very effective. The key trick here is the randomization introduced by extremely random trees proves to be effective in predicting more events accurately. We call these models - Boosted random forests (\gls{BRF}) and Boost extremely random trees (\gls{BXT}). The algorithms are tested on the $H \rightarrow \tau^{+} \tau^{-}$ binary classification task. The next chapter presents some results and useful insights, it also draws comparisons with the leading solutions to the problem. 

\begin{sidewaysfigure}
\includegraphics[scale=0.6]{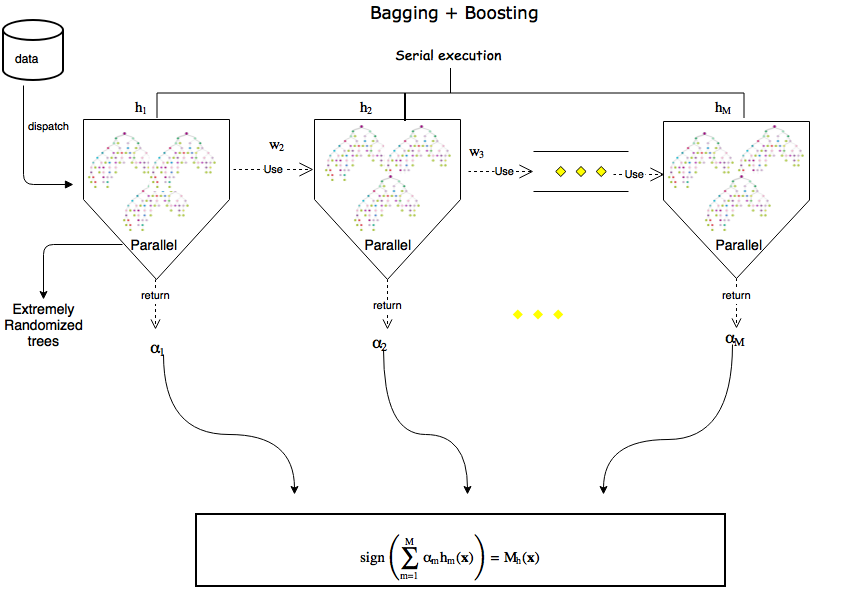}
\caption{Unified bagging + boosting algorithm using extremely random trees}
\end{sidewaysfigure}

\chapter{Results on the ATLAS Higgs dataset}
\label{results}
\section{Exploratory Data Analysis}

\subsection{Data Semantics}
The ATLAS Higgs dataset from the CERN Open data portal has a total of 800K collision events which have been simulated by the high energy ATLAS simulator. The simulator encompasses the current best understanding of the physics underlying the signal and background decays. The signal events mimic properties of the background events as is known to happen in real events. All the events are labelled - either $s$ or $b$ which enable supervised learning techniques to be employed on the data. 

For simplicity, ATLAS uses only three sources of background in creating this particular dataset. Information on the source of each background process is not provided in the dataset. The most abundant background source comes from the decay of the $Z$ boson to two taus. The $Z$ has a mass (91.2 Gev) very close to that of the Higgs (125 Gev) and $Z$ decay occurs much more frequently than the Higgs decay. 

The events are broken up into three categories to facilitate the learning task. The learning task comprises of training, cross-validation and testing and there can be no leakage of data from one group to the other. For this thesis we use the ATLAS proposed division of the data. The density of events in each of the three individual sets in highlighted in table \ref{break}.
This is to aid in analysis of the data and prevent users from splitting the data into training and test sets in a way that distorts the density of signal events by over-sampling or under-sampling.  

\begin{table}[H]
\begin{center}
\begin{tabular}{l|c|c|c}
Dataset & Total events & Background $|\mathcal{B}|$ &  Signal $|\mathcal{S}|$ \\
\toprule
Training & 250,000 & 164,333 & 85,667 \\
Cross Validation & 100,000 & 65,975 & 34,025\\
Testing & 450,000 & 296,317 & 153,683\\

\end{tabular}
\caption{ATLAS Breakdown of datasets for the learning task}
\label{break}
\end{center}
\end{table}
Some of the feature values in the dataset have the tag -999.0. They cannot exactly be termed as missing data since data is structurally absent in these cases and cannot be computed. For instance, data with PRI$\_$jet$\_$num = 0 means the collision event did not produce any jets, this implies that such events also do not have any of the features based on jet properties. In a similar fashion, there are features which are structurally undefined for several events. The fraction of events undefined for each feature and class are summarized in the table below: 
 
\begin{table}[H]
\begin{tabular}{l|c|c|c}
Feature & Background undefined & Signal undefined & Total undefined\\
\toprule
PRI$\_$tau$\_$pt &  - & - & \rdelim\}{10}{20pt}[\parbox{12.5mm}{None}] \\
PRI$\_$tau$\_$eta & - & - & \\
PRI$\_$tau$\_$phi & - & - & \\ 
PRI$\_$lep$\_$pt & - & - & \\
PRI$\_$lep$\_$eta & - & - & \\
PRI$\_$lep$\_$phi & - & - & \\
PRI$\_$met & - & - & \\
PRI$\_$met$\_$sumet & - & - & \\
PRI$\_$met$\_$phi & - & - & \\
PRI$\_$jet$\_$num & - & - & \\
\midrule
PRI$\_$jet$\_$leading$\_$pt & \rdelim\}{3}{20pt}[30$\%$] & \rdelim\}{3}{20pt}[10$\%$] & \rdelim\}{3}{20pt}[40$\%$]\\ 
PRI$\_$jet$\_$leading$\_$eta & & & \\
PRI$\_$jet$\_$leading$\_$phi & & & \\
\midrule
PRI$\_$jet$\_$subleading$\_$pt & \rdelim\}{3}{20pt}[50$\%$] & \rdelim\}{3}{20pt}[21$\%$]  & \rdelim\}{3}{20pt}[71$\%$] \\
PRI$\_$jet$\_$subleading$\_$eta & & & \\
PRI$\_$jet$\_$subleading$\_$phi &  & & \\
\midrule
PRI$\_$jet$\_$all$\_$pt & - & - & - \\
\midrule
DER$\_$mass$\_$MMC & 14$\%$ & 1$\%$ & 15$\%$ \\
\midrule
DER$\_$mass$\_$transverse & - & - & \rdelim\}{3}{20pt}[\parbox{12.5mm}{None}] \\
DER$\_$mass$\_$vis & - & - & \\
DER$\_$pt$\_$h & - & - & \\
\midrule
DER$\_$deltaeta$\_$jet$\_$jet & \rdelim\}{4}{20pt}[50$\%$] & \rdelim\}{4}{20pt}[21$\%$]  & \rdelim\}{4}{20pt}[71$\%$] \\
DER$\_$mass$\_$jet$\_$jet & & & \\
DER$\_$prodeta$\_$jet$\_$jet & & & \\
DER$\_$lep$\_$eta$\_$centrality & & & \\
\midrule
DER$\_$deltar$\_$tau$\_$lep & & & \rdelim\}{5}{20pt}[\parbox{12.5mm}{None}] \\
DER$\_$pt$\_$tot & - & - & \\
DER$\_$sum$\_$pt & - & - & \\
DER$\_$pt$\_$ratio$\_$lep$\_$tau & - & - & \\
DER$\_$met$\_$phi$\_$centrality & - & - & \\
\midrule
\end{tabular}
\caption{Fraction of undefined values in the ATLAS Higgs dataset}
\label{missing_values}
\end{table}

It is important to point out that almost all undefined features are related to \texttt{PRI$\_$jet$\_$num} = 0 except \texttt{DER$\_$mass$\_$MMC}. From the table we gauge that 40$\%$ of events had no jets and hence no leading jet properties and 71$\%$ of events had less than 2 jets (atleast 2 jets are needed for subleading jet properties). Ignoring events which have an undefined tag in any of the features gets rid of a large number of events. For instance, more than 62$\%$ of the signal events lack atleast one feature, hence we need to work with events that have missing values.  

Traditionally, missing values are imputed with either the mean or the median to bring the data-point back to into the normal range of values. However, such imputation creates artificial peaks in the distribution of the features and it is important to verify it does not distort classification performance. For features with skewed distributions, the median is a better imputer than mean as the median is more robust to outliers. 

Each event in the dataset has an importance weight $w_{i}$, they were introduced in chapter \ref{formal} (see eq. \ref{wratio}). The weight for a signal event is on average 300 times smaller than the weight for a background event. Due to this, weights cannot be used as features as their magnitude can predict signal and background events perfectly. However, they do form a part of the training process but not as a non-traditional feature.

\begin{figure}[H]
\includegraphics[scale=0.5]{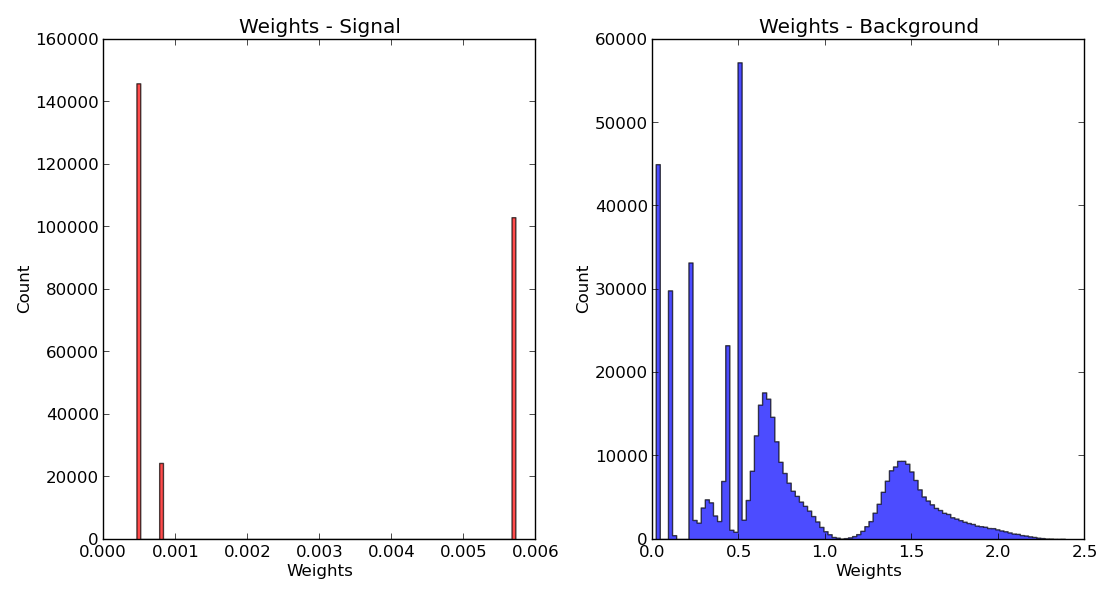}
\caption{Distribution of weights of the signal and background class}
\label{weightsd}
\end{figure}

Fig. \ref{weightsd} depicts the distribution of weights across the whole dataset for signal and background classes. The peaks in the background weight distribution correspond to the individual sources of background events used by the simulator. In the weight distribution for signals we see 3 distinct spikes, they indicate Higgs production by 3 unique mechanisms. Apart from the weights there is no information in the dataset to indicate the specific signal mechanism or background source. 

\subsection{Features}

\begin{figure}
\includegraphics[width=\textwidth]{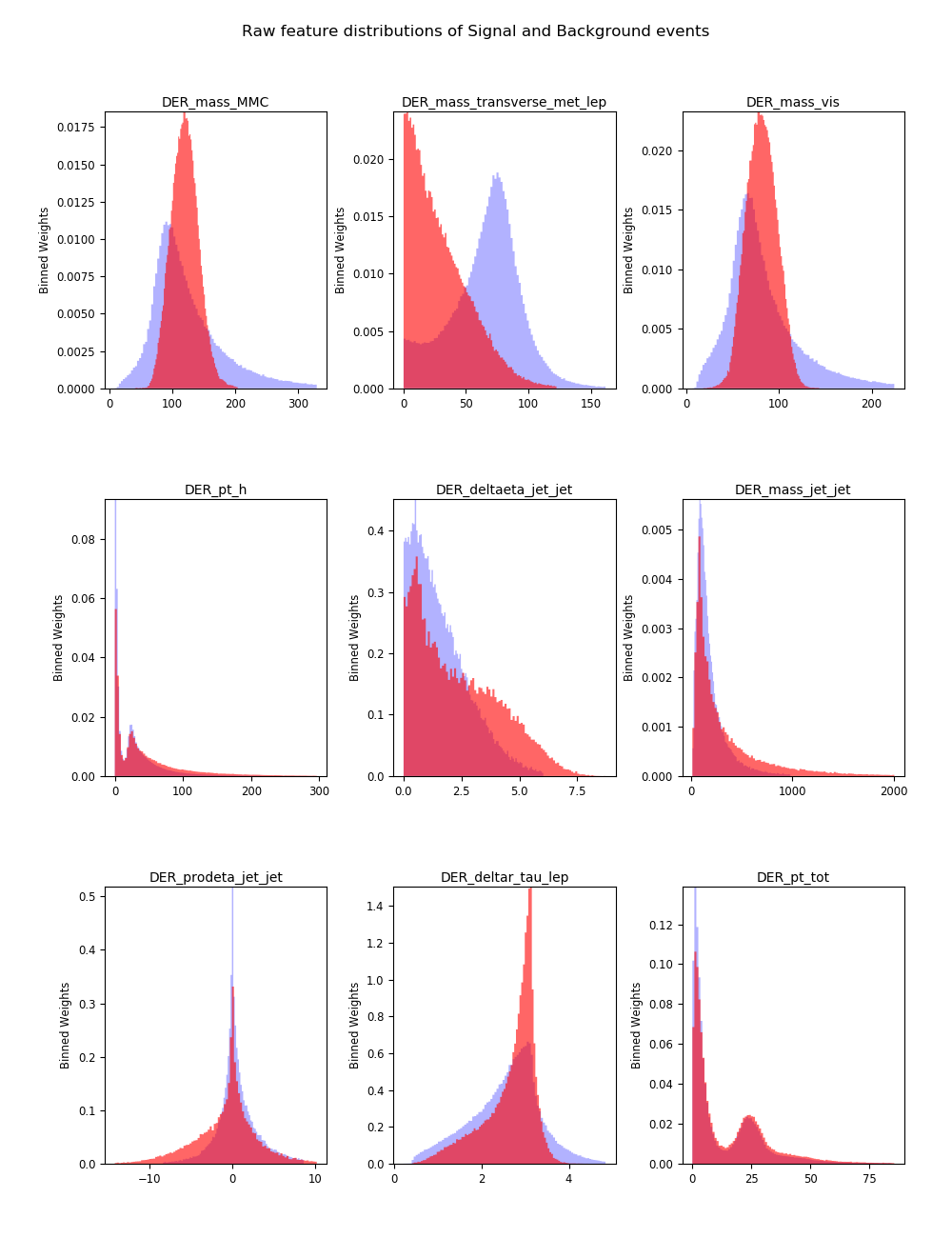}
\caption[Raw feature distributions]{Raw feature distributions of signal (red) and background (blue) events depicting values on their true scale, further in order to make the histogram more representative we sum importance weights $w_{i}$ in each bin rather than count.}
\label{unscaled_features1}
\end{figure}

\begin{figure}
\includegraphics[width=\textwidth]{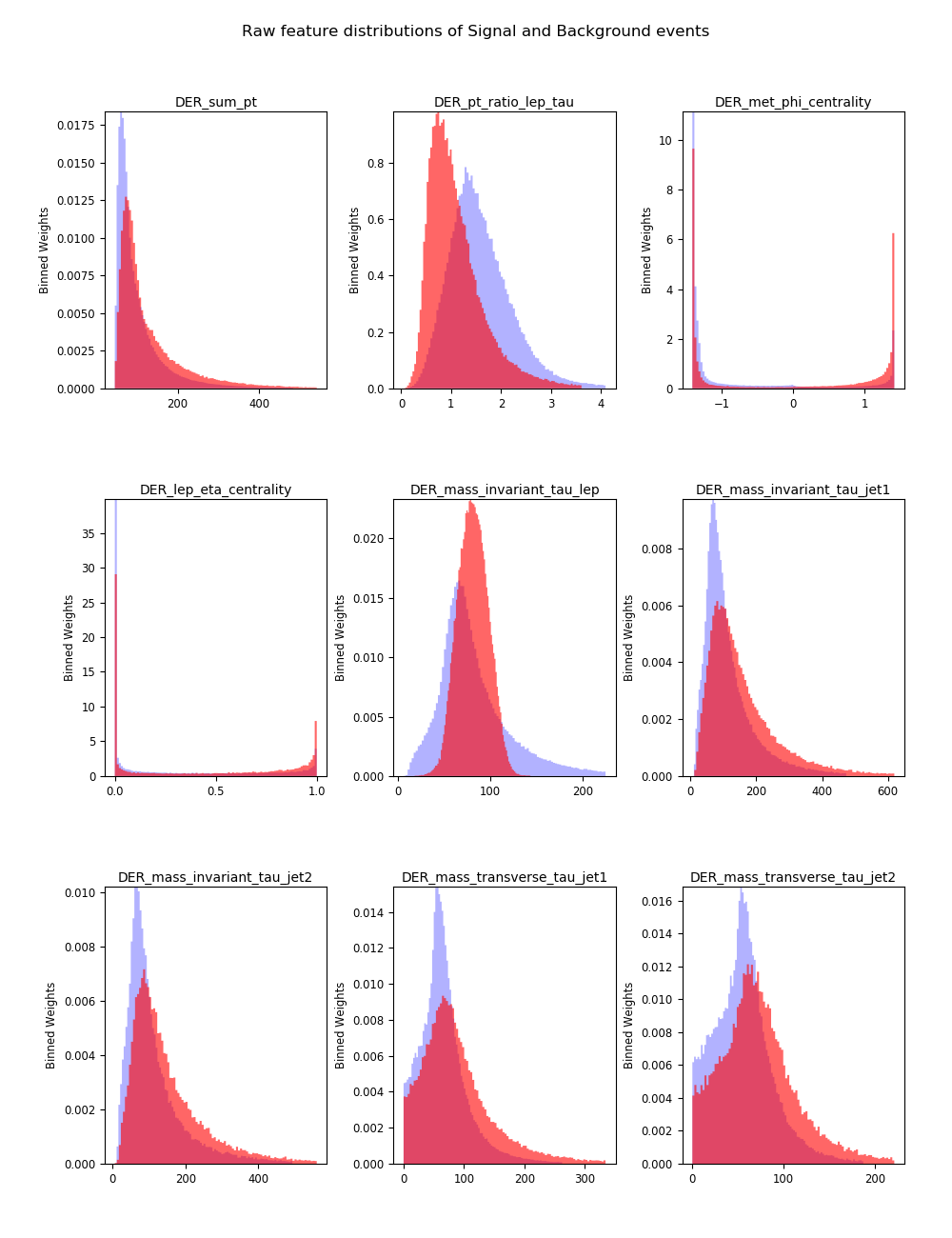}
\caption[Raw feature distributions]{Raw feature distributions of signal (red) and background (blue) events depicting values on their true scale, further in order to make the histogram more representative we sum importance weights $w_{i}$ in each bin rather than count.}
\label{unscaled_features2}
\end{figure}

ATLAS provides 30 features for each signal and background event in the dataset. These features were described section \ref{features}. The features are partitioned into primary and derived features. The derived features are more discriminatory than the primary features as the former represent quantities calibrated by physicists while the latter are raw momenta of particles observed in the decay channel. In fig. \ref{unscaled_features1} and \ref{unscaled_features2} we depict the distributions of derived features split by signal and background events.

\subsection{Class overlap}

\begin{figure}[ht]
\includegraphics[width=\textwidth]{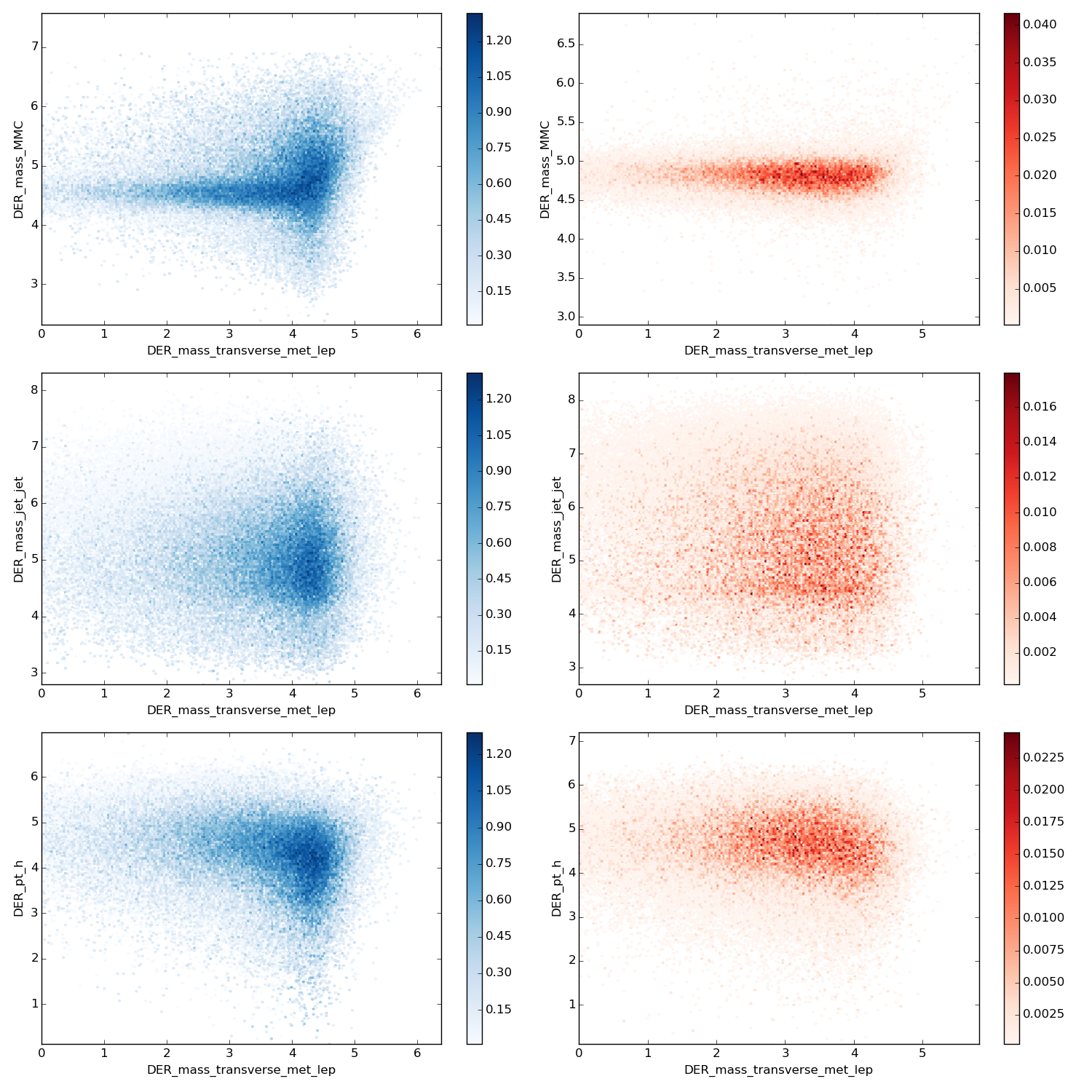}
\caption{Density of weights of signal and background expressed in 2d space.}
\label{scatter1}
\end{figure}

\begin{figure}[ht]
\includegraphics[width=\textwidth]{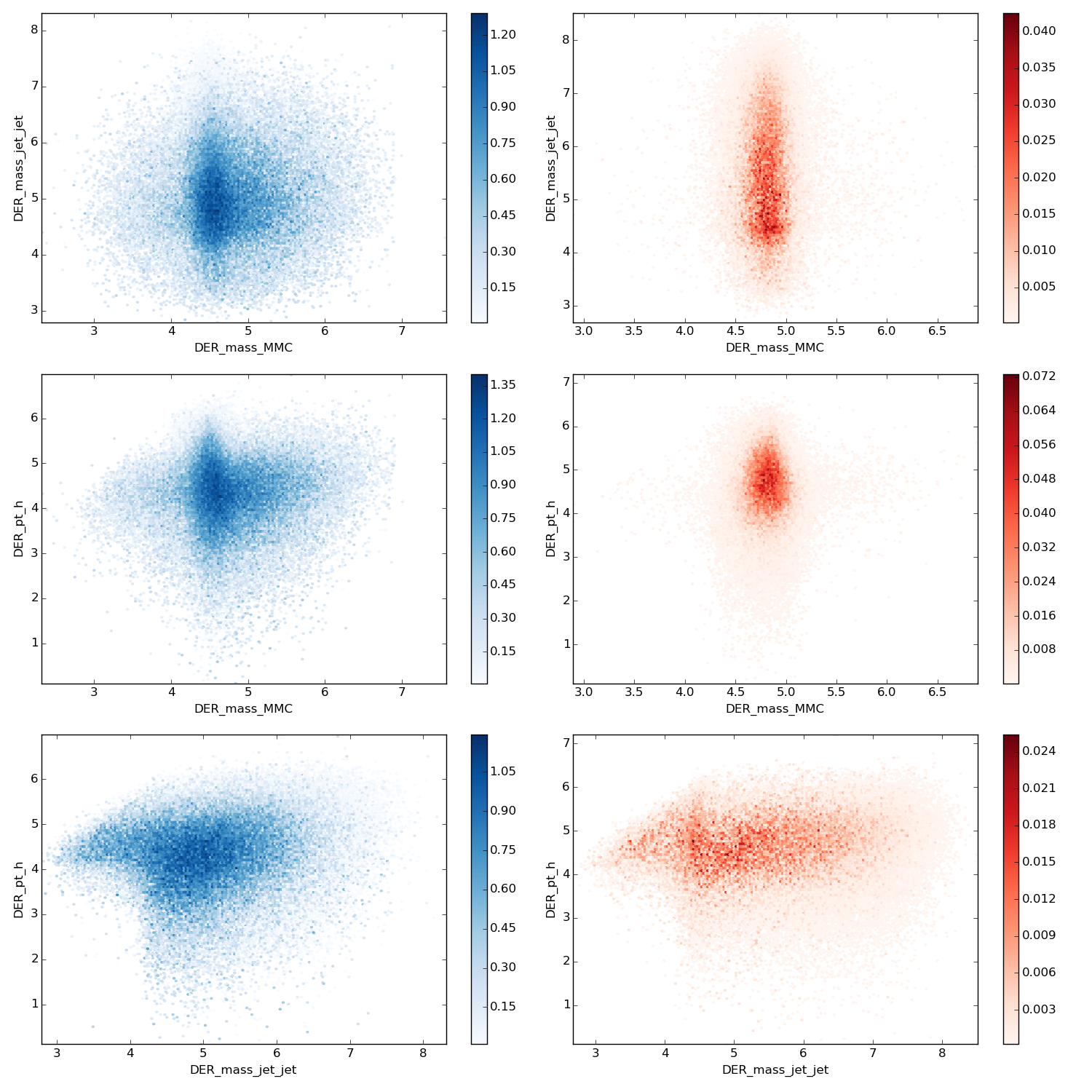}
\caption{Density of weights of signal and background expressed in 2d space.}
\label{scatter2}
\end{figure}

Fig. \ref{scatter1} and \ref{scatter2} provide a sense of the difficulty of the task of classifying background and signal events. Not only do the classes predominantly overlap, the densities of background and signal events captured by the importance weights $w_{i}$ attached to each event show that the signal is sparse and embedded in the background.  The features in the plots represented are from the more discriminatory derived features. 

\section{Tree ensembles}

The results presented in this section make use of the bagging principle used in random forests (RF) and extremely random trees (ET). The aim of this section is to establish a baseline performance for bagging methods using the primitive tree learner and provide some insight into the variations between random forests and extremely random trees. 

\begin{figure}[h]
\centering
\includegraphics[scale=0.5]{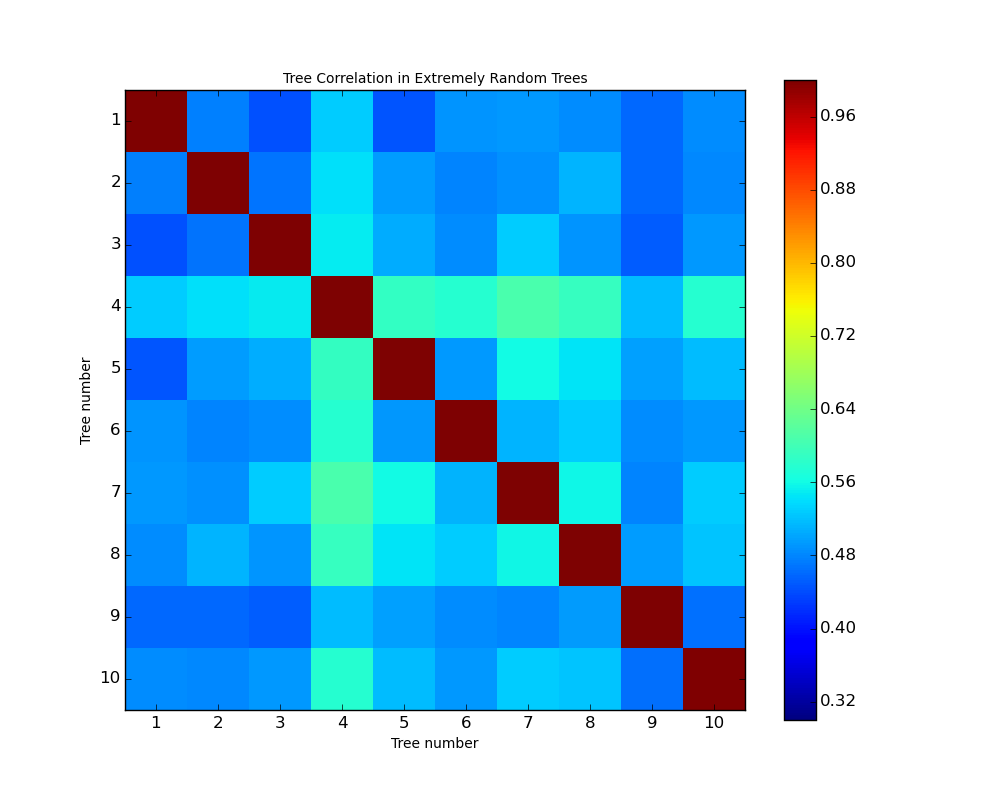}
\caption{Correlation between tree outputs under the ET tree ensemble.}
\label{tree_correlation}
\end{figure}

\begin{figure}[h]
\centering
\includegraphics[scale=0.5]{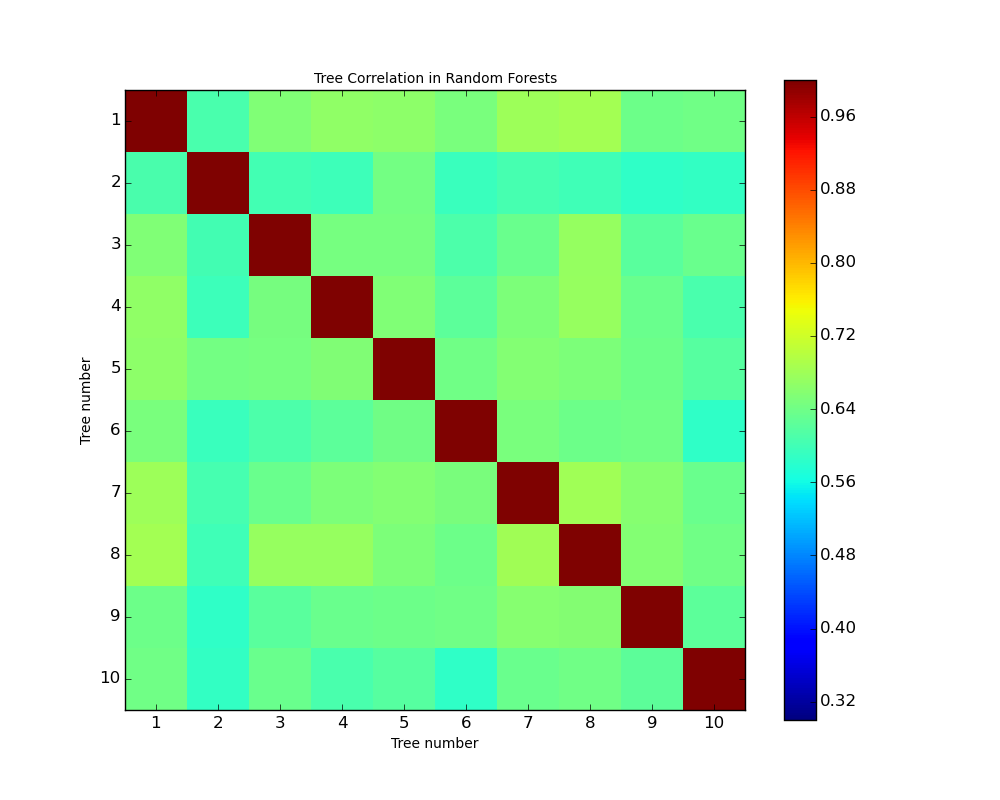}
\caption{Correlation between tree outputs under the RF tree ensemble.}
\label{et_tree_correlation}
\end{figure}

Intrinsic to the construction of the tree learner is the splitting principle, each tree in a RF uses a splitting rule based on a metric that scores all possible splits in a collection of features and then chooses the best split. On the other hand, extremely random trees do not use a splitting rule. They construct trees by choosing the best split from a set of random splits (see algorithm \ref{randomsplits}). This attribute of extremely random trees make them less accurate than random forests but much faster. Fig. \ref{tree_correlation} and \ref{et_tree_correlation} depicts the correlation between trees in a RF and ET model. ET trees are visibly less correlated than RF trees. The diagonal entries indicate the correlation of a tree output with itself. 

\begin{figure}[h]
\includegraphics[scale=0.6]{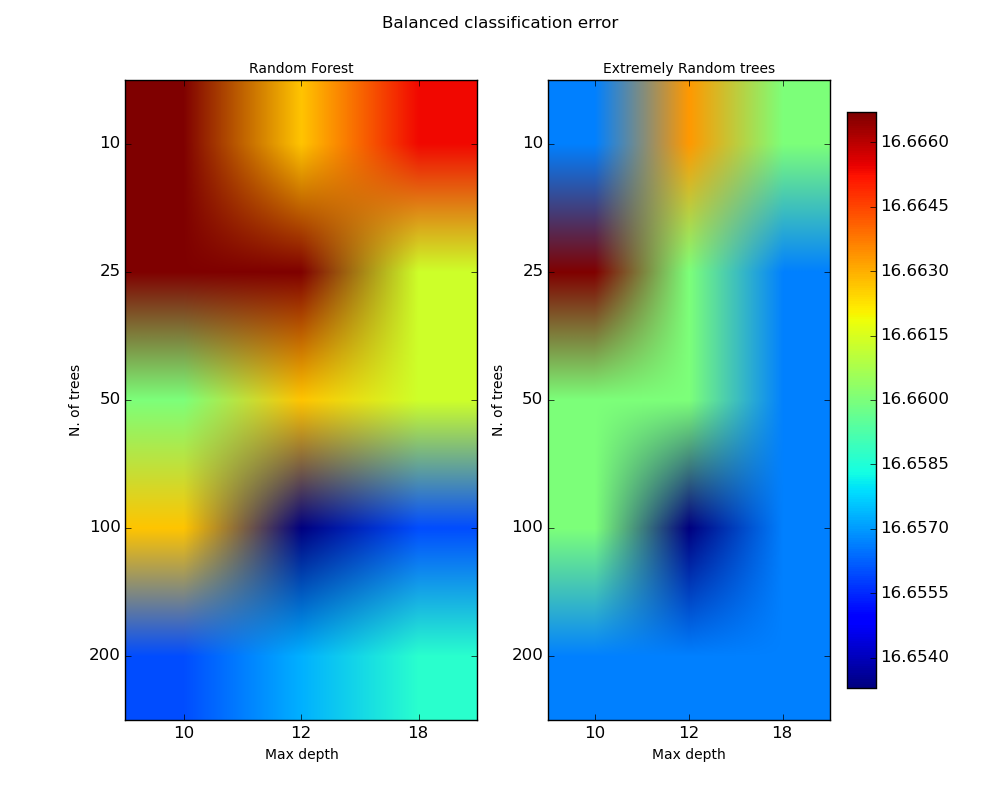}
\caption{Density of balanced classification error for tree ensembles - Max depth vs. number of trees}
\label{tree_num_features}
\end{figure}

Figure \ref{tree_num_features} shows the density of the balanced classification error on a grid of two parameters. The parameters here are the \textit{Max depth} and \textit{Max features}. The former refers to the maximum allowed depth of a tree and the latter refers to the maximum features considered to split on at each node. The magnitude of change in the balanced classification error is small due to the magnitude of the balanced weights (see fig. \ref{bweights}). A 0.001 difference in the balanced classification error amounts to on average 100 misclassified events. An interesting observation is that the tree ensembles saturate at 100 primitive learners and increasing the number of learners to 200 leads to a deterioration in error. 

The significance curves of RF and ET models are depicted in section \ref{sig} below. The next section presents results on the workings and behaviour of two meta-ensembles and highlights some important differences which make one more markedly suited to the classification task than the other. 
  
\section{Meta Ensembles}

In this section we analyse the results of two models on the Higgs dataset - boosted random forests (BRF) and boosted extremely random trees (BXT). As the name suggests each model works on the principle of boosting, but uses different base learners. As opposed to simple BDTs which boost a single decision tree, in BRFs and BXTs we boost a tree ensemble. An automated parameter search for the right parameters of such models is hard due to the combinatorial explosion in the parameter space. Instead the approach followed here is to look for convergence of the ensemble effect with respect to the AMS ($\sigma$).  We take our cue from the optimization in the earlier section where we analysed the performance and behaviour of random forests (RF) and extremely randomized trees (ET). These models are the base learners in the meta ensemble.  

\subsection{De-construction of Boosting}

\begin{figure}[h]
\includegraphics[width=\textwidth]{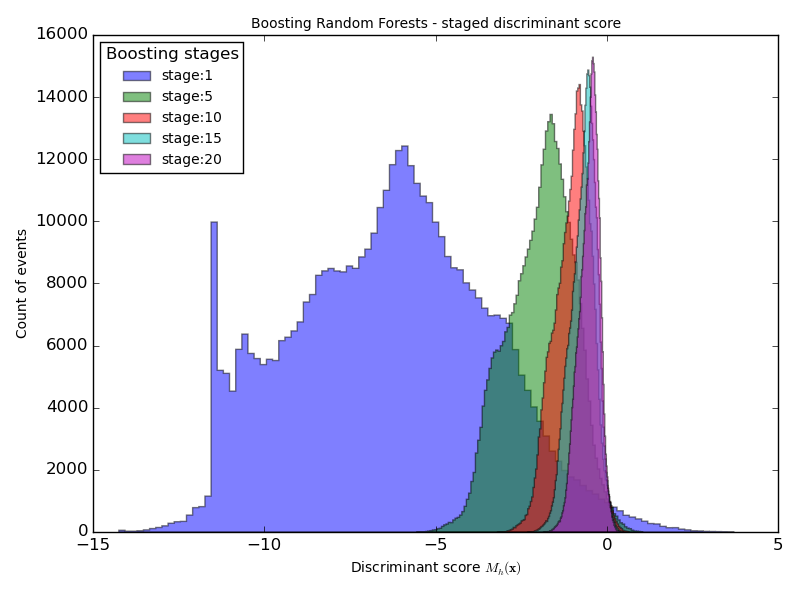}
\includegraphics[width=\textwidth]{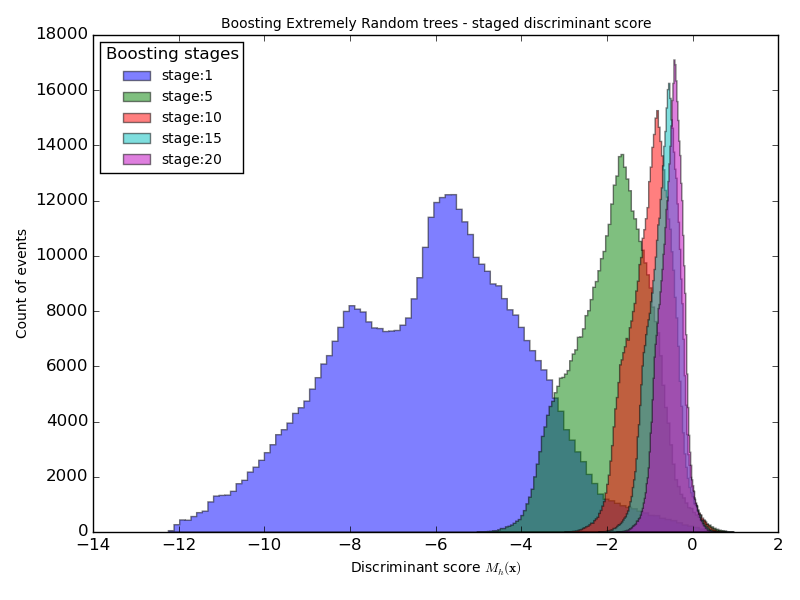}
\caption[Evolution of discriminant score in BRF and BXT.]{Apart from the noticeable spike in the distribution of BRFs which results from a large number of samples getting an identical score, we can notice that the BRFs in stage one have a thicker right tail as opposed to the one of BXTs.}
\label{staged}
\end{figure}

In this section we analyse the effect that boosting has on the discriminant scoring function. The shape of the distribution of the discriminant score is critical to the selection of a pure selection region. Distributions which are spread out and fat-tailed indicate low confidence in scoring and tend to give poorer AMS scores. In figures \ref{staged} we observe the evolution of the discriminant score distribution at successive levels of boosting. We can notice how the incremental gains from boosting fall as the number of stages increases. This indicates some level of saturation in the models ability to predict more samples correctly than it already has in the previous stages. The distributions growing more peaked indicate greater confidence in the predicted scores.  
 
\begin{figure}
\includegraphics[width=\textwidth]{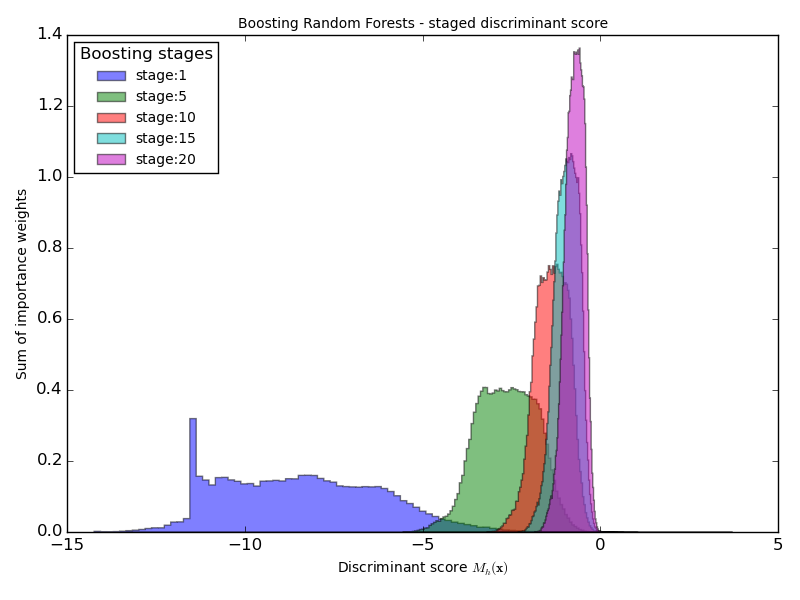}
\includegraphics[width=\textwidth]{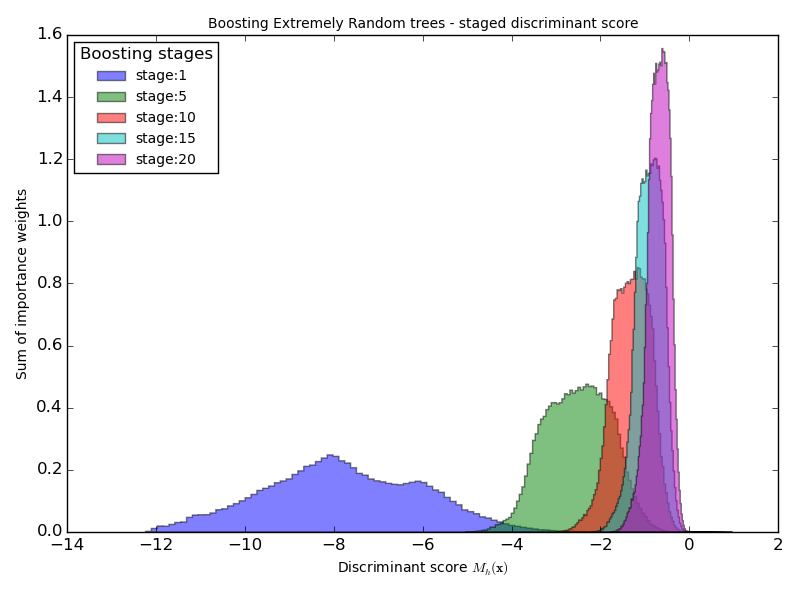}
\caption[Evolution of weighted discriminant score in BRF and BXT]{The spike in events in the BRF score distribution is translated to a spike in the weights in the same region. At stage 20, the BXT model has a visibly higher peak than the BRF model.}
\label{staged_weighted}
\end{figure}

Since the AMS objective is tied to importance weights of each sample, it is often instructive to look at the staged distributions by contribution of weight of events in each bin (of the histogram). In figures \ref{staged_weighted} we represent this weighted score distribution in stages.  

\begin{sidewaysfigure}
\includegraphics[scale=0.7]{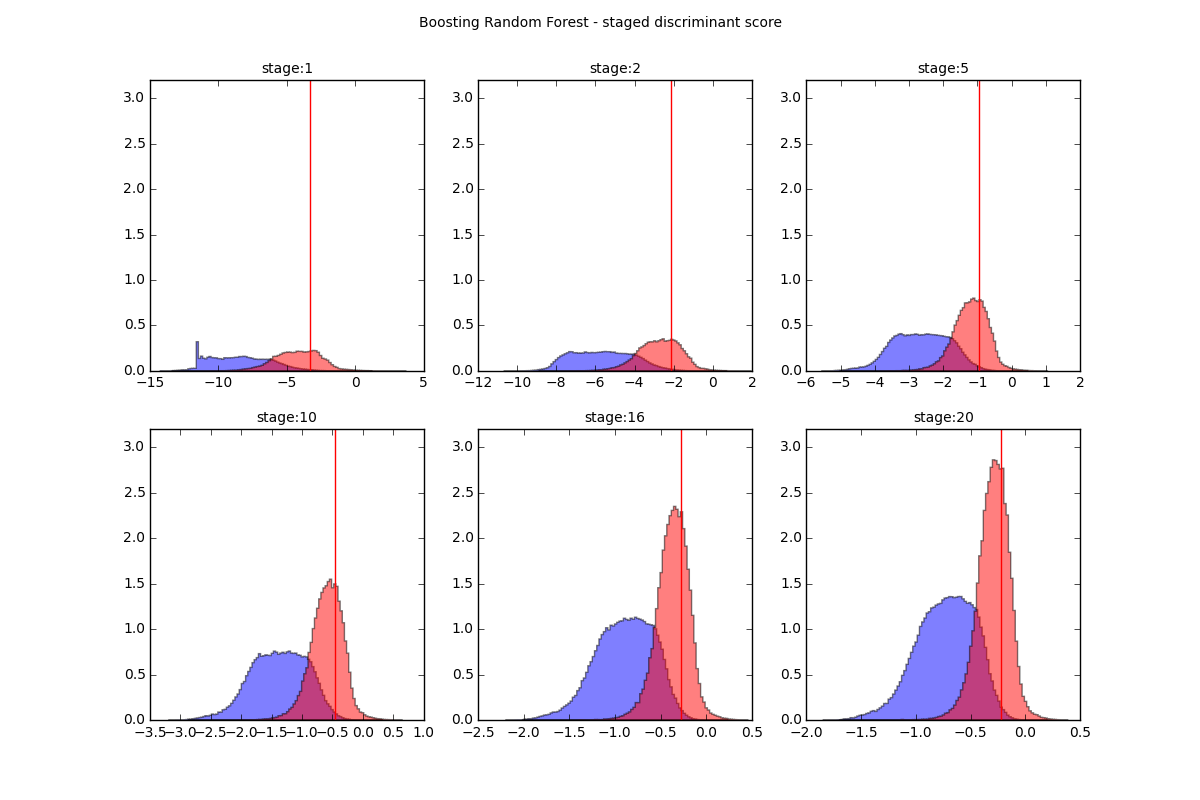}
\caption[Evolution of weighted discriminant score in BRF.]{Event which lie on the right hand side of this cut determine the selection region.}  
\label{brf_staged_weighted_bi}
\end{sidewaysfigure}

\begin{sidewaysfigure}
\includegraphics[scale=0.7]{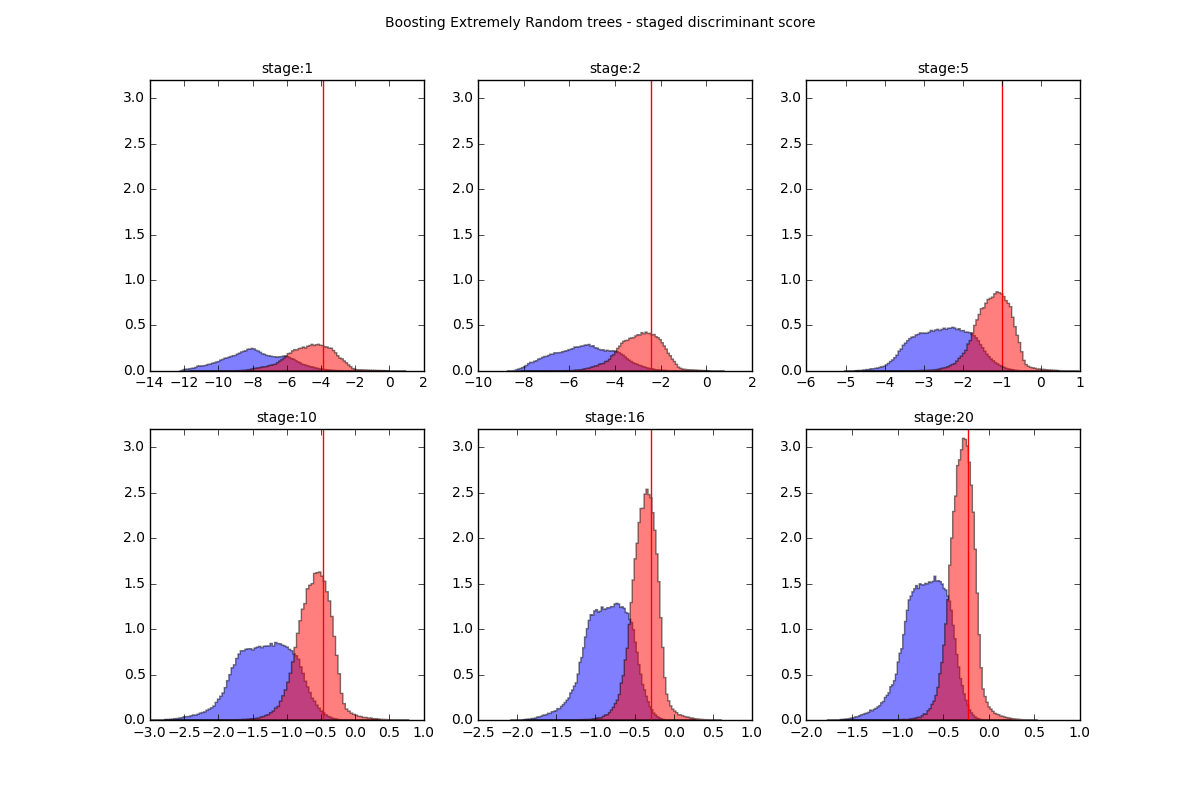}
\caption{Evolution of weighted discriminant score in BXT.}
\label{bxt_staged_weighted_bi}
\end{sidewaysfigure}

In figures \ref{brf_staged_weighted_bi} and \ref{bxt_staged_weighted_bi} we observe the weighted distribution of the discriminant score bifurcated by class. This is useful as it points to the region of overlap in the scores of signal and background events. The vertical red line indicates a typical cut (at the 85th percentile) that determines the selection region. The background events which lie on the right hand side of this cut are the false positives and directly impact the AMS $(\sigma)$. Boosting at a fundamental level targets the overlap between signal and background events successively. This is contrast to complex single stage classifiers that do not target the overlap issue per se, their effectiveness is mainly tied to the creation of complex decision boundaries in higher dimensional spaces. 

\begin{figure}
\includegraphics[width=\textwidth]{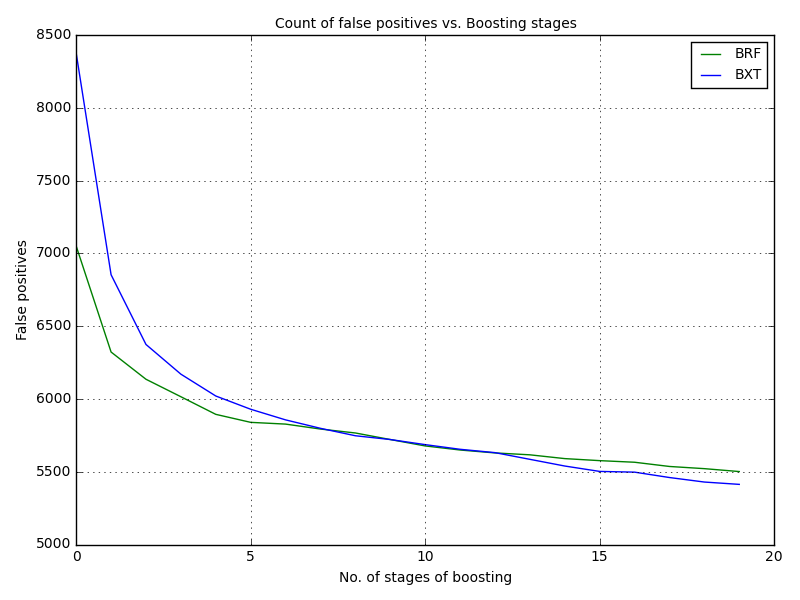}
\includegraphics[width=\textwidth]{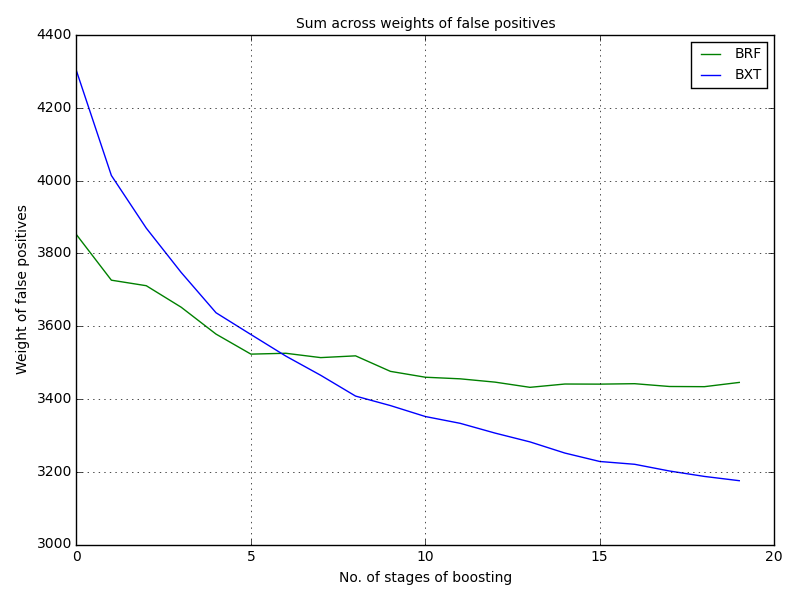}
\caption{Count and weights of false positives in the selection region for classifiers - BRF and BXT}
\label{fp_evolve}
\end{figure}

From figures \ref{brf_staged_weighted_bi} and \ref{bxt_staged_weighted_bi} we see the narrowing of the width of the discriminant function but it is not clear if the actual count of false positives is going down in successive stages. Figure \ref{fp_evolve} depicts this information for both the models. Both the count and sum across weights of false positives ($\hat{b}$) drop with number of stages. In terms of count, it seems like both the models perform equally well however, in terms of the importance weights the models diverge after around 5 rounds of boosting. While the BRF is more effective in earlier stages at weeding out the false positives, its effectiveness saturates rapidly after round 5. The BXT model exhibits a slow and consistent decline in the weighted false positives. After 20 rounds, it is evident that BXT the selection region has fewer false positives than the BRF model.   

\subsection{Tree correlations}

\begin{figure}
\includegraphics[width=\textwidth]{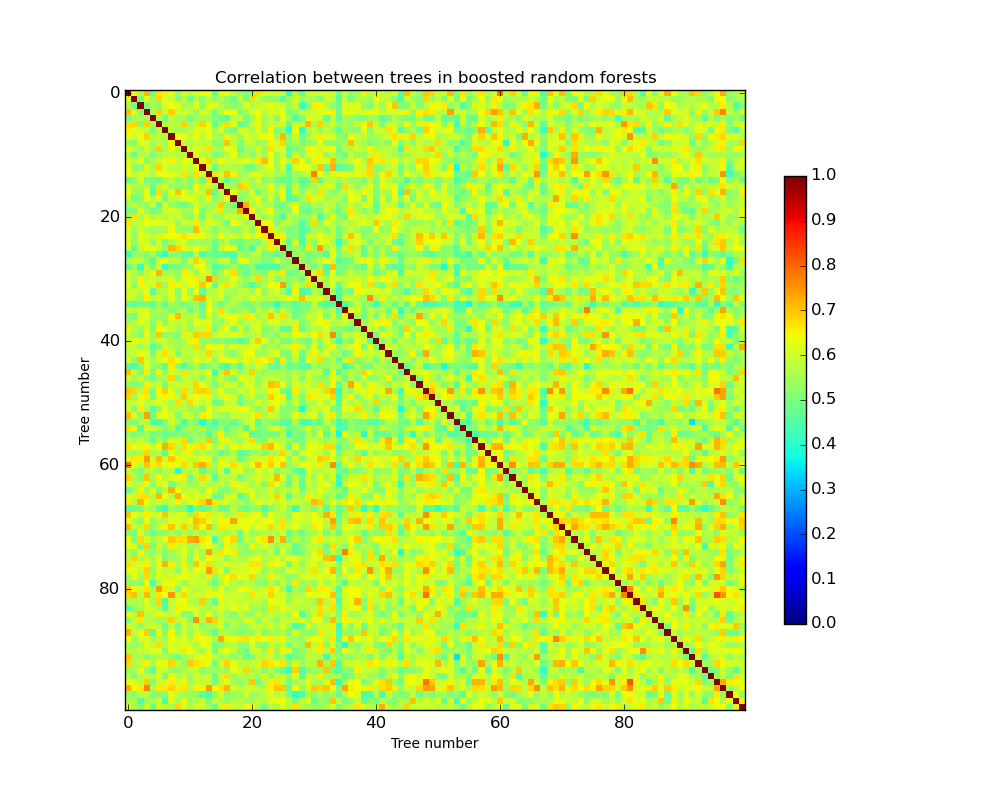}
\includegraphics[width=\textwidth]{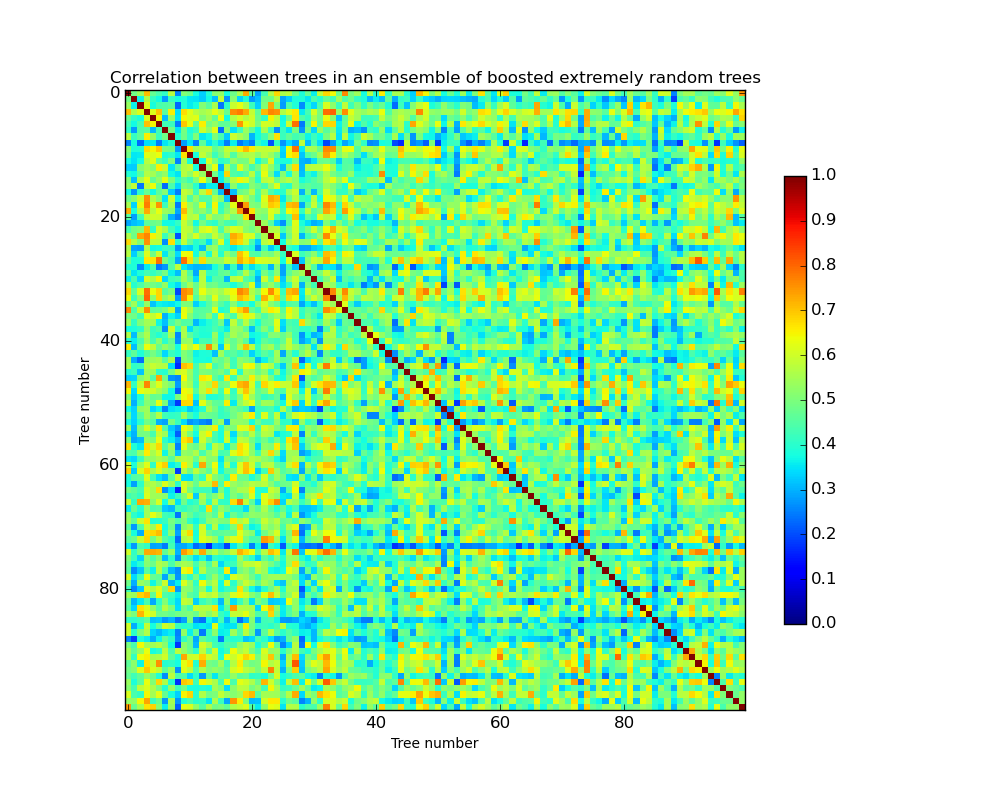}
\caption{Correlation between trees in BRF and BXT.}
\label{boosted_correlations}
\end{figure}

Correlation between trees is fundamental to the effectiveness of the BXT model. Figure \ref{boosted_correlations} depicts the correlation between 100 primitive trees picked at random from 
the 2000 trees that comprise the final stage BRF and BXT model (20 stages of boosting $\times$ 100 trees in each tree ensemble). The higher level of de-correlation between the outputs of BXT versus the BRF indicate two critical aspects:

\begin{itemize}
\item BXT creates more diverse tree ensembles than BRF, diversity in the context of classification means that each tree ensemble is good at predicting different events. Diversity in the ensembles gives the model access to predicting a larger number of points correctly. The BRF trees are more correlated in their predictions, saturating their predictive ability.
\item BXT is able to benefit from more rounds of boosting than BRF, as we saw in fig. \ref{fp_evolve}. The main reason for this is that the BRF model is a strong predictor and boosting on strong predictors is not as beneficial as boosting on moderately weak predictors. 
\end{itemize} 
 
\section{Significance Curves \texorpdfstring{$(\sigma)$}{}}
\label{sig}

For the tree ensembles proposed in this study, the AMS ($\sigma$) is the definitive measure of performance. In this section we summarize the results of the AMS objective introduced in detail in chapter \ref{formal}. The AMS ($\sigma$) is computed using the events in the selection region of a classifier. The AMS is essentially the ratio $\hat{s}/\sqrt{\hat{b}}$ where $\hat{s}$ and $\hat{b}$ are the sum across importance weights of signal and background events in the selection region (see \ref{unbiased}). The exact mathematical form used in the figures in this section is, 

\begin{equation}
\sqrt{2((\hat{s} + \hat{b})\ln(1 + \frac{\hat{s}}{\hat{b}})-\hat{s})} 
\end{equation} 

The AMS ($\sigma$) appears on the $y$ axis. In order to facilitate comparison with the leading solution to the $H \rightarrow \tau{+}\tau^{-}$ classification problem presented in the ATLAS Higgs dataset, each plot has a horizontal red line which indicates the leading published score on this dataset.\footnote{This was the winning score on the ATLAS Kaggle machine learning competition held in 2014.} 

\subsection{Single Ensembles}

\begin{figure}
\includegraphics[width=\textwidth]{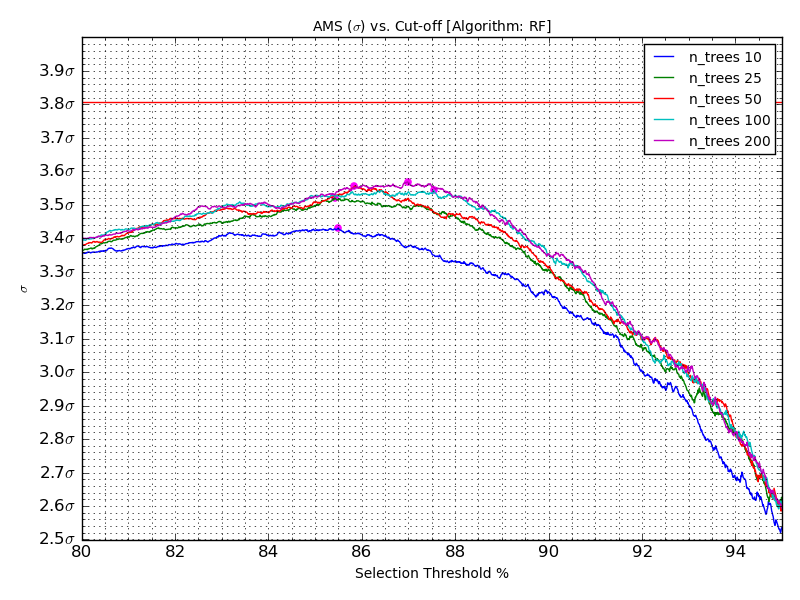}
\includegraphics[width=\textwidth]{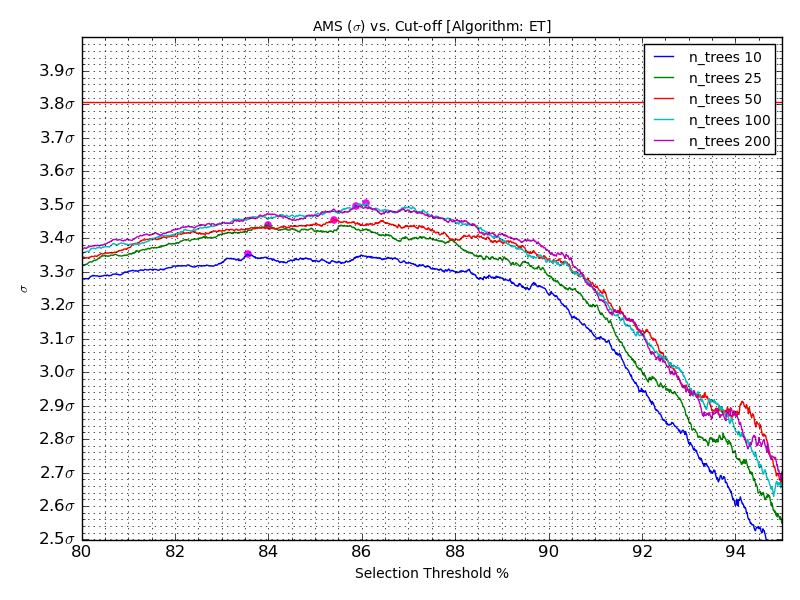}
\caption[AMS curves corresponding to increasing number of trees.]{We can observe the progression of curves in the ET model more clearly than in the RF model.}
\label{cluster}
\end{figure}

Fig. \ref{cluster} depicts the evolving AMS curves for a random forest ensemble (RF) and the extremely random trees ensemble (ET). The curves are evolving with respect to the number of primitive tree learners in the ensemble. This figure depicts the `bagging' effect in the context of inseparable classes. Aggregating a large number of primitive learners improves the overall performance. Averaging predictions reduces the variance of the predictions but increases the bias, the overall classification error is attributed to both - variance and bias. This was discussed in chapter \ref{performance}. The effect we see here is that the improvement in error due to reduction in variance is large enough to offset the increase in bias due to averaging.  

\begin{figure}
\includegraphics[width=\textwidth]{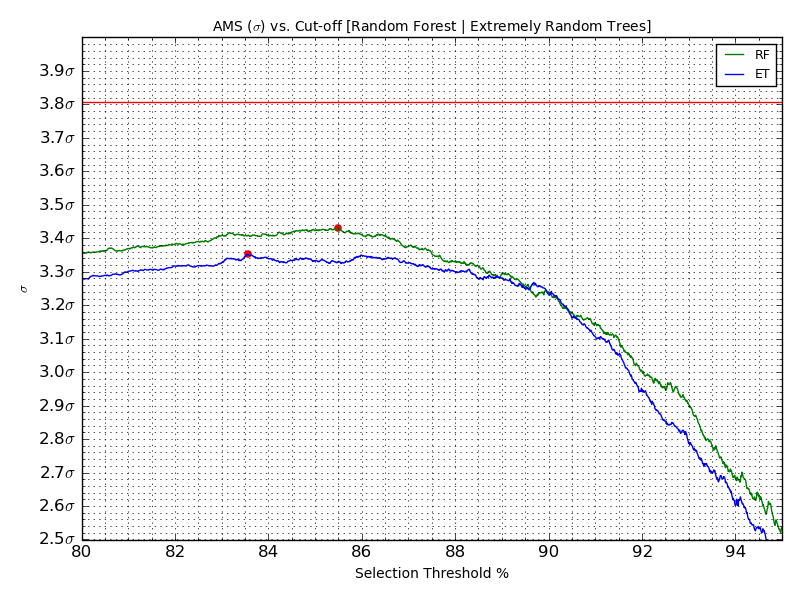}
\includegraphics[width=\textwidth]{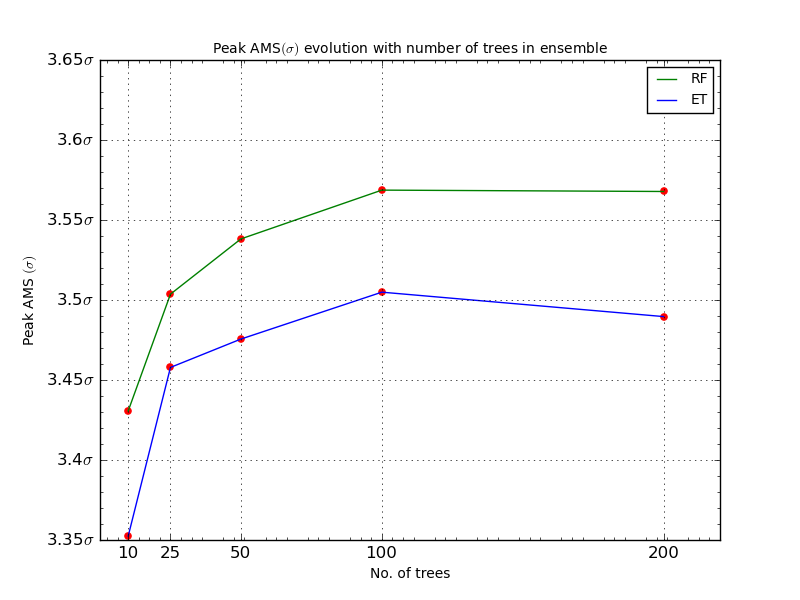}
\caption{Comparitive performance of AMS on the RF and ET model}
\label{ams_et_rf}
\end{figure}

It is important to note that the individual RF model is slightly better than the individual ET model at all levels. In order to put the significance curves of each algorithm into perspective, the figure \ref{ams_et_rf} shows the evolution of the best significance ($\sigma$) with respect to the number of trees in the ensemble for each algorithm. It is essentially a plot of the magenta dots ({\color{magenta}\small{$\bullet$}}) on each  of the curves in fig. \ref{cluster}.

\subsection{Boosted Ensembles}

\begin{figure}
\includegraphics[width=\textwidth]{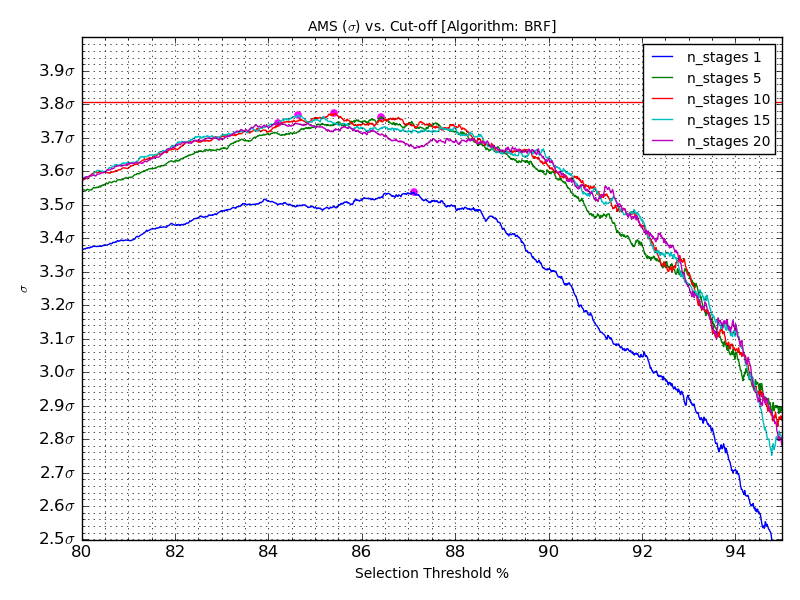}
\includegraphics[width=\textwidth]{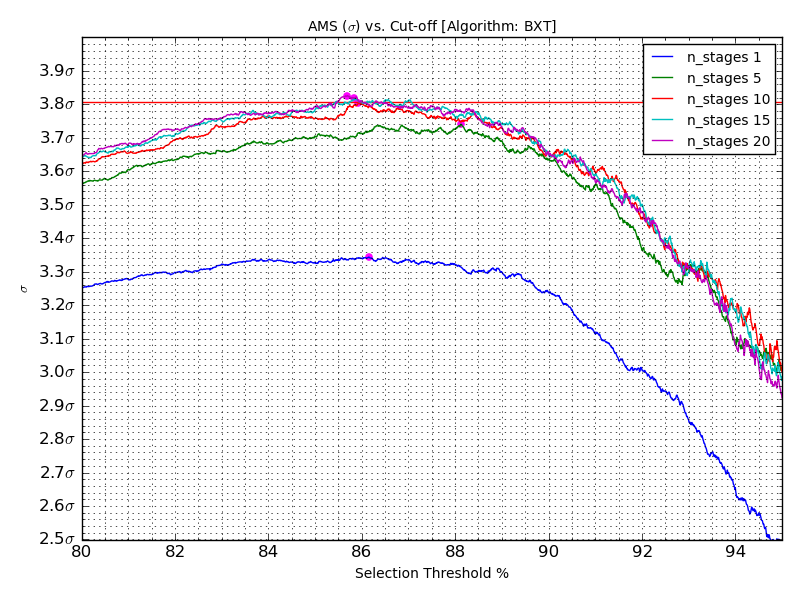}
\caption{AMS curves at $n$ stages of boosting}
\label{ams_boosting}
\end{figure}

This section depicts the significance curves of boosted ensembles, each tree ensemble goes through several stages of boosting where the misclassified samples from the previous stages are over weighted. While a single extremely random trees classifier (ET) falls short of performing as well as a random forest, it is clear that in the boosted incarnation of the ensembles, BXTs steal an edge over the forests. In fig. \ref{ams_boosting} we observe a marked progression of curves in the BXT model in response to boosting while in the BRF model, the effect is more muted. 

\begin{figure}
\includegraphics[width=\textwidth]{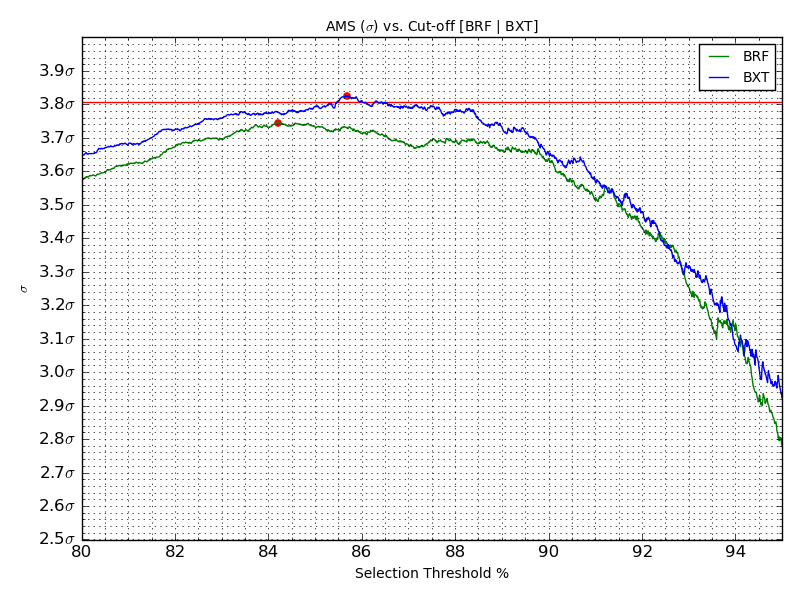}
\includegraphics[width=\textwidth]{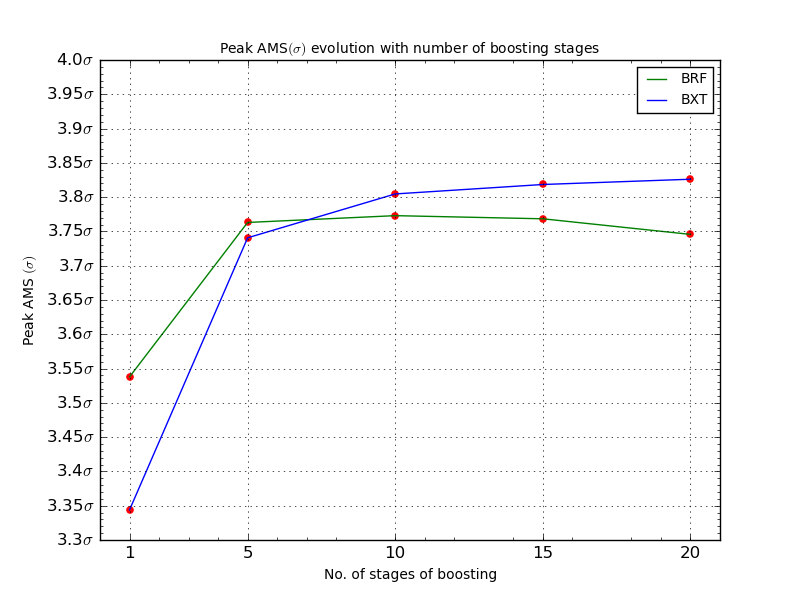}
\caption{Best AMS curves for boosted forests (BRF) and boosted extremely randomized trees (BXT).}
\label{ams_brf_bxt}
\end{figure}

Figure \ref{ams_brf_bxt} depicts the AMS performance at the end of 20 rounds of boosting tree ensembles. BXT with extremely random trees as primitive learners give a consistently better AMS ($\sigma$) value across a range of selection thresholds. In order to facilitate a comparison between the two models, they use the same number of primitive trees (no. of trees = 100) in each stage of boosting. The AMS for the BXT model is comparable to the leading solution on this dataset. The evolution of the peak AMS gives further insight into the behaviour of the models. The peak AMS of the BRF model saturates within a few stages of boosting indicating that further stages do not add any value to the model, the same behaviour is seen in the BXT model but at a much later stage of boosting.   

\section{Summary of Tree Ensembles}

The table \ref{treeams} summarizes the performances of tree ensembles. All results reported are based on the test dataset of 450K events. 

\begin{center}
\begin{table}[H]
\resizebox{\textwidth}{!}{
\begin{tabular}{l|c|c|c|c}
Learning Algorithm & Selection Region & False Positives & AMS at 85th percentile ($\sigma$) & Primitive Learners\\
\toprule
Random Forest (RF)  & 67632 & 7132 & 3.42294$\sigma$ & 100\\
Extremely Random Trees (ET)  & 67766 & 8106 & 3.33157$\sigma$ & 100 \\
Boosted Random Forests (BRF)  & 66789 & 5501& 3.73311$\sigma$ & 20 $\times$ 100\\
Boosted Extremely Random Trees (BXT)  & 66876 & 5413 & 3.79361$\sigma$ & 20 $\times$ 100\\
\end{tabular}}
\caption{Performance metrics of tree ensembles}
\label{treeams}
\end{table}
\end{center}

The BXT ensemble shows it is possible to leverage the performance of blazingly fast primitive tree learners for complex classification tasks where the data is inseparable and overlapping. Primitive trees have a computational edge over other classifiers albeit they give simple decision boundaries compared to more powerful classifiers like kernel support vector machine and neural nets. However, in several problems trees have shown to claw back the loss in accuracy if enough of them are harnessed in a meta approach. The next section discusses the computational elements of tree ensembles and draws some perspectives by comparing to the leading solution.  

\section{Computing}

The architecture of the meta algorithm BXT imposes some level of serial execution. Since the training occurs iteratively in independent stages the pipeline for training and testing essentially work in a serial fashion. 

Within each stage of training we use a randomized forest as a base learner, this algorithm works on the principle of bagging which is an apt candidate for parallel execution. In bagging, the training of each bootstrap sample is conducted independently, hence they can be generated in parallel. 

The size and dimensionality of the dataset were relatively tractable and did not prove to be computational bottlenecks. The full dataset comprised of 800K samples and 30 features, 5 redundant features were dropped and 9 additional features were added during preprocessing leaving the original size relatively unchanged. An advantage of tree learners is their training speed relative to models like neural networks which are much slower to train. Python provides an efficient version of the CART algortihm in its \texttt{scikit-learn} package which was used to train the primitive learners in the ensemble.  
 
The table \ref{treetimes} summarizes the training speed of the algorithms alongside their implementation details.

\begin{table}[H]
\resizebox{\textwidth}{!}{
\begin{tabular}{l|c|c|c|c|l}
Learning Algorithm & Training speed & Learning Stages & No. of Trees & Tree Construction & Machine specs \\
\toprule
Single Decision Tree  & 4.77 sec. & 1 & 1 & Serial &  \rdelim\}{6}{20pt}[\parbox{18.5mm}{Intel Xeon (R) CPU \\ 3.10GHz x 4}] \\
Boosted Single decision trees (BDT) & 45.28 sec. & 30 & 1 & Serial  \\
Random Forest (RF) & 70 sec. & 1 & 30 &  Parallel &  \\
Extremely Random trees (ET) & 26 sec. & 1 & 30 &  Parallel & \\
Boosted Random Forests (BRF) & 1025 sec. &  20 & 100 & Parallel &   \\
Boosted Extremely Random Trees (BXT) & 983 sec. & 20 & 100 & Parallel &  \\
\end{tabular}}
\caption{Runtime of tree ensembles on a single CPU 4-core machine}
\label{treetimes}
\end{table}

In the table we choose the same number of trees for the bagged models (RF, ET) as the number of stages in BDT to facilitate a comparison between the two from a computational viewpoint. The computational load of training a single tree 30 times (one time per stage for BDT) is identical to training 30 trees one time (bagged, RF and ET). Extremely random trees are faster than random forests as they by pass the search for the best split at each node of the primitive tree learner. The computational advantage coupled with a consistently higher AMS ($\sigma$) curve point to the effectiveness of the meta algorithm in the Higgs classification task. Gabor Melis (the owner of the lead solution) trained his ensemble of neural networks on a GTX Titan GPU where training a single 3-layer network took 10 minutes. As a comparison the solution provided in this thesis runs on a 4-core CPU machine and the entire model trains in under 20 mins. The boosted extremely random trees (BXT) algorithm proposed in this thesis achieves an AMS score of 3.79361$\sigma$ at the 85th percentile cut-off.

\begin{figure}
\includegraphics[width=\textwidth]{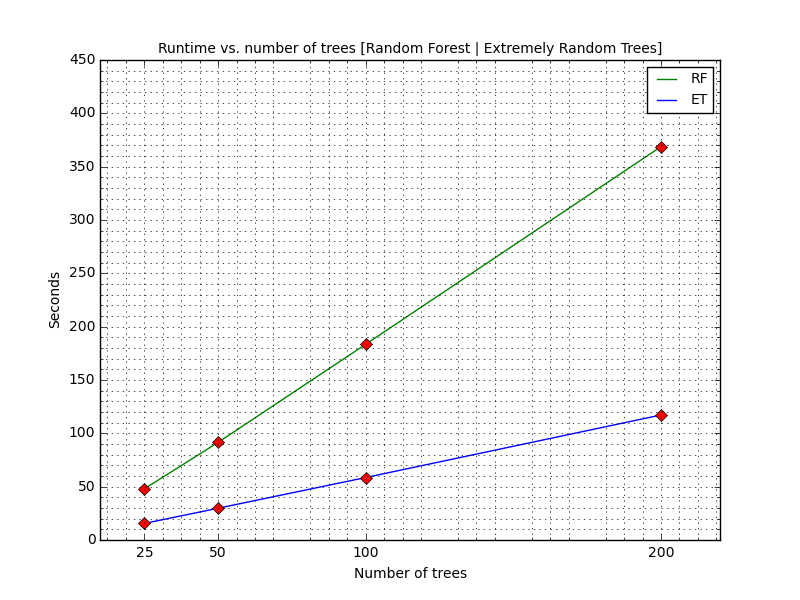}
\caption{Training Speed}
\label{runtime}
\end{figure}

Fig. \ref{runtime} depicts the training speed of ETs versus RFs, they grow linearly in the number of trees in the ensemble.  

\section{Complexity}

In this section we briefly summarise the computational complexity of the learning algorithms used in this thesis. We present this information using the `Big $\mathcal{O}$ notation' which in this context describes how the algorithm scales to changes in the input size. Mathematically, it provides an upper bound for the growth of a function ignoring lower order terms. 

\begin{equation}
f(n) = n^{4} + 4n^{3} + n^{2} + 7 = \mathcal{O}(n^{4}) \textrm{ as } n \rightarrow \infty 
\end{equation}

\subsection{Single Decision Tree}

The cost of constructing a binary tree using $n$ samples and 1 feature vector is $\mathcal{O}(nlog_{2}(n))$. In the presence of $s$ features, splitting at each node requires searching through $\mathcal{O}(s)$ features to find the best split, this amounts to a total complexity of $\mathcal{O}(sn\log_{2}(n))$. In order to query a single data point (pass an unseen data point down a tree until it reaches a leaf), the complexity is $\mathcal{O}(\log_{2}(n))$. 

\subsection{Tree ensemble}

In a random forest, where we specify the number of trees, say $M$ trees, the complexity is $\mathcal{O}(Msn\log_{2}(n))$. However, this is not exact as in a random forest, each split considers a random subset of features and this random selection of features adds an additional overhead.  

The $\mathcal{O}(Msn\log_{2}(n))$ complexity is worst-case assuming that the depth of the tree is going to be $\mathcal{O}(\log_{2}(n))$. In most cases the stopping criterion causes the tree to terminate much before it attains its maximum depth and hence this upper bound is rarely reached. For instance, if we set a criterion for the minimum number of samples required for a node split to be 500, nodes with less than 500 samples would convert to leaves and stop growing on that branch. Rules like these prune the unbounded growth of a tree to its maximum achievable depth, they serve as over fitting control. Trees fitted to a training dataset that are allowed to grow to their maximum depth are rarely useful for prediction on new samples as they over fit the training dataset. If the depth of each tree is specified as a parameter before training, the tree stops growing when it achieves the pre-specified depth. In this case the worst case complexity is simplified to $\mathcal{O}(Msnd)$ where $d$ is the pre-specified tree depth.

In order to query a single data point on a random forest with $M$ trees of depth $\log_{2}(n)$, the complexity is $\mathcal{O}(M\log_{2}(n))$. With $M$ trees of depth $d$, it becomes $\mathcal{O}(Md)$.

\subsection{Boosted ensemble}

The complexity of a boosting algorithm where the complexity of the underlying base learner, say $T$ is $\mathcal{O}(T)$ is $\mathcal{O}(NT)$ where $N$ is the number of rounds of boosting. The dependence is trivially linear as in each stage the cost is $\mathcal{O}(T)$ and with $N$ stages, it is $\mathcal{O}(NT)$. Where T is a random forest with $M$ trees, the cost of boosting $N$ random forests is $\mathcal{O}(NMsn\log_{2}(n))$. The cost of querying a $N$ stage boosting algorithm with a random forest (of $M$ trees) as base learner is $\mathcal{O}(NM\log_{2}(n))$.


\chapter{Conclusion}
\label{concl}

\section{Contributions of this study}

\begin{enumerate}

\item This thesis proposes a meta-learning algorithm that adopts a unique approach of utilizing principles that counterbalance each other - boosting and bagging. Bagging generates more robust models at the expense of bias and boosting transforms weak learners into stronger ones. Boosting a traditional forest of trees did not give competitive results as the learners it created were not diverse enough. BXT in essence tries to induce diversity in its outputs at the expense of accuracy by making them less deterministic (random splits for a random choice of features). This makes the extremely random tree learner a sub-optimal stand alone model which very often is less competitive to the traditional random forest \cite{xrf}. However, the randomization trick makes the extremely random tree a good choice for boosting as it is in a position to harness the weight updates in boosting for several stages until it saturates. This thesis demonstrates the effectiveness of a boosted tree ensemble versus the traditional approach in HEP of boosting single decision trees. 

\item This thesis provides a systematic study of a challenging classification problem presented in the ATLAS Higgs dataset from the perspective of a machine learner. By providing a model which achieves a competitive AMS score it sets an effective performance threshold achievable by using the basic tree learner in an innovative re-incarnation. It also provides a threshold against which more sophisticated models can be rated and compared. This is in keeping with the Occam's razor approach which states simpler models should be preferred over more complex ones if their performances are similar. In conducting this study it has come to light that some of the solutions proposed for achieving high scores on this dataset were powerful but superfluous \cite{melis}. Similar scores are achievable using much simpler architectures, one such has been described in this thesis. 

\end{enumerate}

\begin{figure}
\includegraphics[scale=0.6]{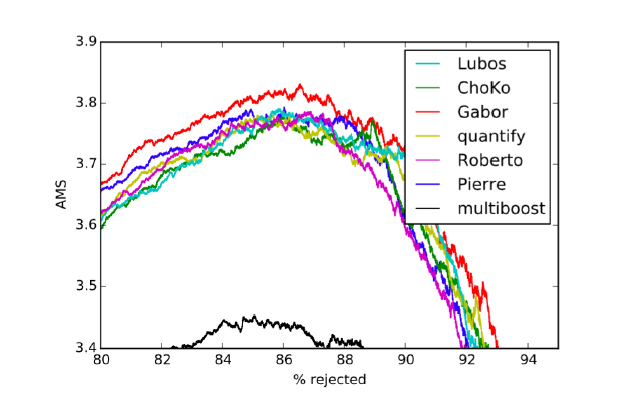}
\caption{Leading Solutions to the Kaggle ATLAS Higgs machine learning challenge. The challenge used the same dataset as the one used in this thesis.}
\label{win}
\end{figure}

In order to put into perspective the model developed in this thesis, the fig. \ref{win} shows the leading solutions to the ATLAS Higgs dataset. 

\begin{figure}
\includegraphics[scale=0.6]{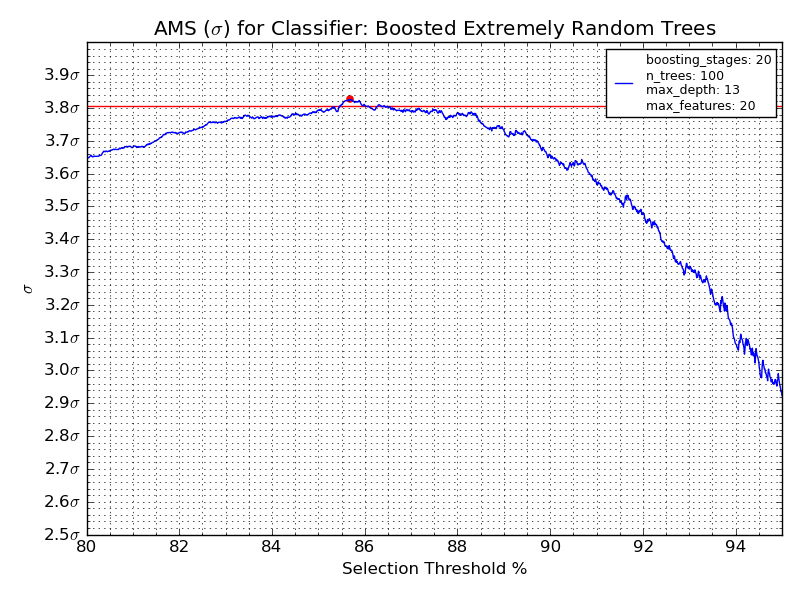}
\caption{BXT solution to the ATLAS Higgs dataset.}
\end{figure}

\section{Related / Future Work}

As any study does, this study also creates some new questions which could be interesting for the HEP community of machine learners at large. They could form the basis of future or related research trajectories. 

\begin{itemize}
\item The work in this thesis provides an innovative way to use a primitive learner like a tree for classification in an extremely challenging setting. The work does not explore changing the way a primitive binary tree learns. One of the fundamental building blocks of tree learning is the splitting criterion, and majority of the studies that focus on applications of tree learning use axis-parallel linear splits .i.e they split on a single feature at each node. Geometrically, these univariate splits at each node can be thought of generating space partitioning hyperplanes parallel to the feature axis. Splitting criterion that generate oblique splits by using a linear combination of attribute values, could perform very differently to the primitive tree with linear splits. They generate polygonal partitions in feature space instead of rectangular ones.  There is very little research on such trees and hardly any software packages offer implement construction of such trees \cite{oblique}. 
\item Bagging and boosting can be combined in various ways, the approach in this thesis using primitive trees is one example, and is in no way the final word on the topic \cite{combining, combining2}. 
\item Deriving theoretical performance bounds on learning from primitive learners like binary trees. As we have seen, additional classification power can be extracted from trees by using enough of them and re-combining them. How far they can be stretched in the context of a specific problem is established through experimentation, by looking for convergence in one or more performance criteria. The analysis present in this thesis is a coarse-grained approach to optimization in the presence of a non-standard performance metric, but can we do better than optimization by hand? 
\end{itemize}

Boosting uses an internal bifurcation procedure to split data into ones that were misclassified and others, in this way it passes information about learning in one stage to the next. The samples that are misclassified in early stages are the ones that lie in the most overlapping regions of the feature space. It is possible to enrich data with this meta-information before training commences or in between the boosting stages. Meta information that captures a spatial attribute of the dataset like degree of overlap for each data point could add classification power to trees which are rule based learners and do not capture this information about the geometric structure of the dataset. This would be a way to combine the power of rule based and distance based learning.  

\begin{sidewaysfigure}
\includegraphics[scale=0.75]{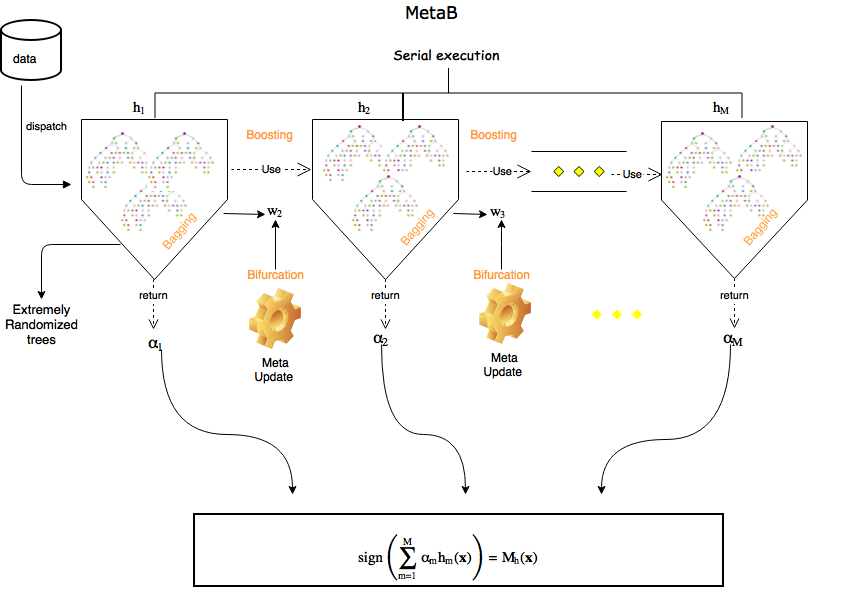}
\caption{Architecture of a proposed algorithm that could combine three meta learning techniques - bagging, boosting and bifurcation in a unified framework}
\end{sidewaysfigure}

\section{Concluding Remarks}

\subsection{On Methodology}

The methodology used in this thesis provides a unified framework for two of the most successful ensemble learning techniques - boosting and bagging. In doing so, it shows that it is possible to achieve performance gains from a carefully fine-tuned combination of the two. The randomization trick used in the BXT model is able to achieve greater diversity in tree outputs while benefiting from boosting for longer than a traditional tree or a random forest. I believe it is this nested combination of randomization within boosting that provides a performance edge in this particular dataset. This is an effective technique to use in the presence of data that is overlapping and ambiguous.

\subsection{On the ATLAS Higgs dataset}

The methodology presented in this thesis presents one approach for tackling the challenges presented in the ATLAS Higgs dataset. The approach suggested here is unique in its study of the problem as it does not rely on the raw classification power of the base learner, rather it relies on the power of randomization and repetition to improve learning in very rudimentary structures like binary trees. This raises the question as to whether enough trees can achieve the same accuracy provided through neural networks which are currently slated to be the most successful learners in this domain. It also brings up a more general question about the links between \textit{learning} and \textit{randomization} in non-linear settings. 

\subsection{On the scientific validity of inference}
Any contemporary science abounds in questions of statistical nature. The current era of science is defined by experiments where the process of scientific discovery hinges upon the correct analysis of scientific data. This is largely because of the ubiquity of data and open access to it. This has also led to the proliferation of modern learning techniques like deep neural networks. The issues involved in designing good standards for scientific inference are challenging and can be thought of as techniques of  meta-analyses (analyses of analyses). Performance assessment is not standardised but is tied to the problem being solved, this creates further challenges. In the Higgs dataset dealt with in this thesis it was not enough to produce classifiers with high accuracy. In fact, that was not the assessment criterion, the assessment criterion was a narrowly defined significance metric, the AMS $\sigma$. What counts as good performance and how to design performance metrics for learning algorithms that cover a vast array of problems? This is a meta field of research which attracts relatively less interest from scientists and practitioners. This is perhaps due to the attraction of objective rules (like a $p$ value $\leq 0.01$ makes for a very significant result rather than $p$ value $\leq 0.03$) and aversion for subjective interpretation. The usage of the Fisherian $p$ value has not changed in the last century and we continue to rely on it in answering questions of great scientific importance in physics, genetics and medicine. The question about whether the $H \rightarrow \tau^{+}\tau^{-}$ decay will ever reach $5\sigma$ is still an open one. With learning techniques rapidly growing in power and sophistication, there are hints that it could. However, even if it does, what does that really mean? 

\bibliographystyle{plain}
\bibliography{Thesis}

\addcontentsline{toc}{part}{\normalsize \appendixname}

\chapter*{Appendix}

\begin{appendices}

\renewcommand{\thesection}{\Alph{section}}

\section{Use of natural units (GeV)}

\label{App}

This section is based on \cite{rm}.

A fundamental equation of special relativity is, $$ E^2 = p^2c^2 + m^2c^4 $$ where $E$ is the energy of the particle, $p$ is its momentum, $m$ is the mass and $c$ is the speed of light. When a particle is at rest its momentum is 0, this gives us Einstein's mass-energy equivalence, $E = mc^2$.  Using the units GeV for Energy, GeV$/c$ for momentum and GeV$/c^2$ for mass we get the equivalence,  $$E^2 = p^2 + m^2 $$ The papers published in the ATLAS and CMS experiment use the notation GeV for mass, energy and momentum. We will follow the same convention.

The momentum $p$ of a particle is actually a 3-dimensional vector $\overrightarrow{p} = (p_{x}, p_{y}, p_{z})$ stating the particle's momentum in 3 directions in 3-d space. For a particle with non-zero mass the momentum of a particle is $\overrightarrow{p} = m\overrightarrow{v}$ where $\overrightarrow{v}$ is the 3-dimensional velocity and $m$ is the mass. The 4-momentum of a particle is defined as $(p_{x}, p_{y}, p_{z}, E)$. This defines the full kinematics of a particle as if we know the particle's momentum and energy we can compute its mass using the relation, $$ m = \sqrt{E^2 - p^2} $$

Similarly, if we know any two quantities out of momentum, mass and energy we can compute the third deterministically by equations of special relativity specified above. 

\clearpage

\end{appendices}

\end{document}